\documentclass[journal]{IEEEtran}

\ifCLASSINFOpdf
  
\else
  
\fi

\usepackage{amsmath}
\usepackage{amsfonts}
\usepackage{url}
\usepackage{graphicx}
\usepackage{booktabs}
\usepackage{color}

\graphicspath{{Figures-PDF/}}

\begin{document}

\title{Thermal Infrared Image Colorization for Nighttime Driving Scenes with Top-Down Guided Attention}

\author{Fuya~Luo,
        Yunhan~Li,
        Guang~Zeng,
        Peng~Peng,
        Gang~Wang,
        and~Yongjie~Li,~\IEEEmembership{Senior~Member,~IEEE}
\thanks{Fuya Luo, Yunhan Li, Guang Zeng, Peng Peng and Yongjie Li are with the MOE Key 
Laboratory for Neuroinformation, the School of Life Science and Tecnology, University of 
Electronic Science and Technology of China, Chengdu 610054, China. E-mail: 
luofuya1993@gmail.com, liyunhan@std.uestc.edu.cn, guangzeng0109@gmail.com, 
pengpanda.uestc@gmail.com, liyj@uestc.edu.cn. (\textit{Corresponding authors: Yongjie Li and Gang Wang.})}
\thanks{Gang Wang is with Center of Brain Sciences, Beijing Institute of Basic Medical 
Sciences, Beijing 100085. E-mail: g\_wang@foxmail.com.}}

\markboth{A Manuscript Submitted to IEEE Transactions on Intelligent Transpotation Systems}
{Shell \MakeLowercase{\textit{et al.}}: Bare Demo of IEEEtran.cls for IEEE Journals}

\maketitle

\begin{abstract}
Benefitting from insensitivity to light and high penetration of foggy environments, 
infrared cameras are widely used for sensing in nighttime traffic scenes. However, the low contrast and 
lack of chromaticity of thermal infrared (TIR) images hinder the human interpretation and 
portability of high-level computer vision algorithms. Colorization to translate a nighttime 
TIR image into a daytime color (NTIR2DC) image may be a promising 
way to facilitate nighttime scene perception. Despite recent 
impressive advances in image translation, 
semantic encoding entanglement and geometric distortion in the NTIR2DC task remain 
under-addressed. Hence, we propose a toP-down attEntion And gRadient aLignment 
based GAN, referred to as PearlGAN. A top-down guided attention module and 
an elaborate attentional loss are first designed to reduce the semantic encoding ambiguity during translation. 
Then, a structured gradient alignment loss is introduced to encourage edge consistency 
between the translated and input images. In addition, pixel-level annotation is carried out on 
a subset of FLIR and KAIST datasets to evaluate the semantic preservation performance of multiple 
translation methods. Furthermore, a new metric is devised to evaluate the geometric 
consistency in the translation process. Extensive experiments demonstrate the superiority of 
the proposed PearlGAN over other image translation methods for the NTIR2DC task. The source 
code and labeled segmentation masks will be available at \url{https://github.com/FuyaLuo/PearlGAN/}.
\end{abstract}

\begin{IEEEkeywords}
Thermal infrared image colorization, image-to-image translation, generative adversarial 
networks, guided attention, nighttime driving scenes perception.
\end{IEEEkeywords}

\IEEEpeerreviewmaketitle

\section{Introduction}

\IEEEPARstart{R}{obust} and reliable all-weather scene perception is essential for intelligent driving 
assistance systems. Nevertheless, imaging devices based on the visible spectrum are extremely sensitive 
to lighting conditions and fail in total darkness, which limits their practicality in severe 
weather and other environments with poor visibiliy. Thermal long-wave infrared (LWIR) 
cameras can capture infrared radiation (7.5--14 $\mu $m) emitted by objects with a temperature 
above absolute zero, which, unlike visible cameras, allows them to image low-light environments without the aid of an 
illumination source. However, thermal infrared (TIR) images captured by LWIR cameras usually 
have low contrast and ambiguous object boundaries. Furthermore, TIR images 
are in shortage of chrominance, which might hinder human interpretation \cite{1999-MT-Cavanillas,1996-MT-Sampson} and subsequent 
transplantation of high-level computer vision algorithms. Therefore, it is of great significance to 
transform a nighttime TIR (NTIR) image into a daytime color (DC) image, which not 
only helps the driver to quickly understand the surrounding environment and sense abnormalities when 
driving at night, but also reduces the annotation burden on NTIR image based computer vision 
tasks by utilizing massively annotated DC image datasets. In this paper, we aim 
to address the problem of NTIR image colorization, which is also called translation from NTIR 
image to DC image (abbreviated as NTIR2DC). 

The aim of NTIR2DC is to map a single-channel TIR image to a 3-channel RGB image and brighten the scene 
without changing the semantics. In general, existing TIR image colorization approaches can be 
categorized into supervised \cite{2018-Qayynm-IBCAST,2018-Berg-CVPRW} and 
unsupervised \cite{2018-Nyberg-ECCV} approaches. Due to the rapid changes in traffic 
scenarios, pixel-level registered infrared-visible image pairs are difficult to acquire, 
which limits the performance of the supervised approaches. In contrast, unsupervised 
image-to-image (I2I) translation offers a potential solution 
to this problem by enforcing the input data distributions of two domains to be similar 
without paired cross-domain images. Recently, the compelling success of Generative Adversarial 
Networks (GAN) \cite{2014-NIPS-Goodfellow} has brought new blood into unsupervised I2I 
translation. To loosen the requirement of pairwise training images, CycleGAN \cite{2017-CVPR-Zhu} 
introduced a cycle consistency loss, which attempts to preserve the original image after a 
cycle of translation and reverse translation. Nyberg \textit{et al.} \cite{2018-Nyberg-ECCV} utilized the CycleGAN model to 
realize unpaired infrared-visible image translation. MUNIT \cite{2018-ECCV-Huang} was 
proposed to improve the diversity of synthetic images. Anoosheh \textit{et al.} \cite{2019-ICRA-Anoosheh} 
utilized a night-to-day image translation model called ToDayGAN to improve image retrieval performance for 
localization tasks. 

\begin{figure}[!t]
\centering
\includegraphics[width=3.45in]{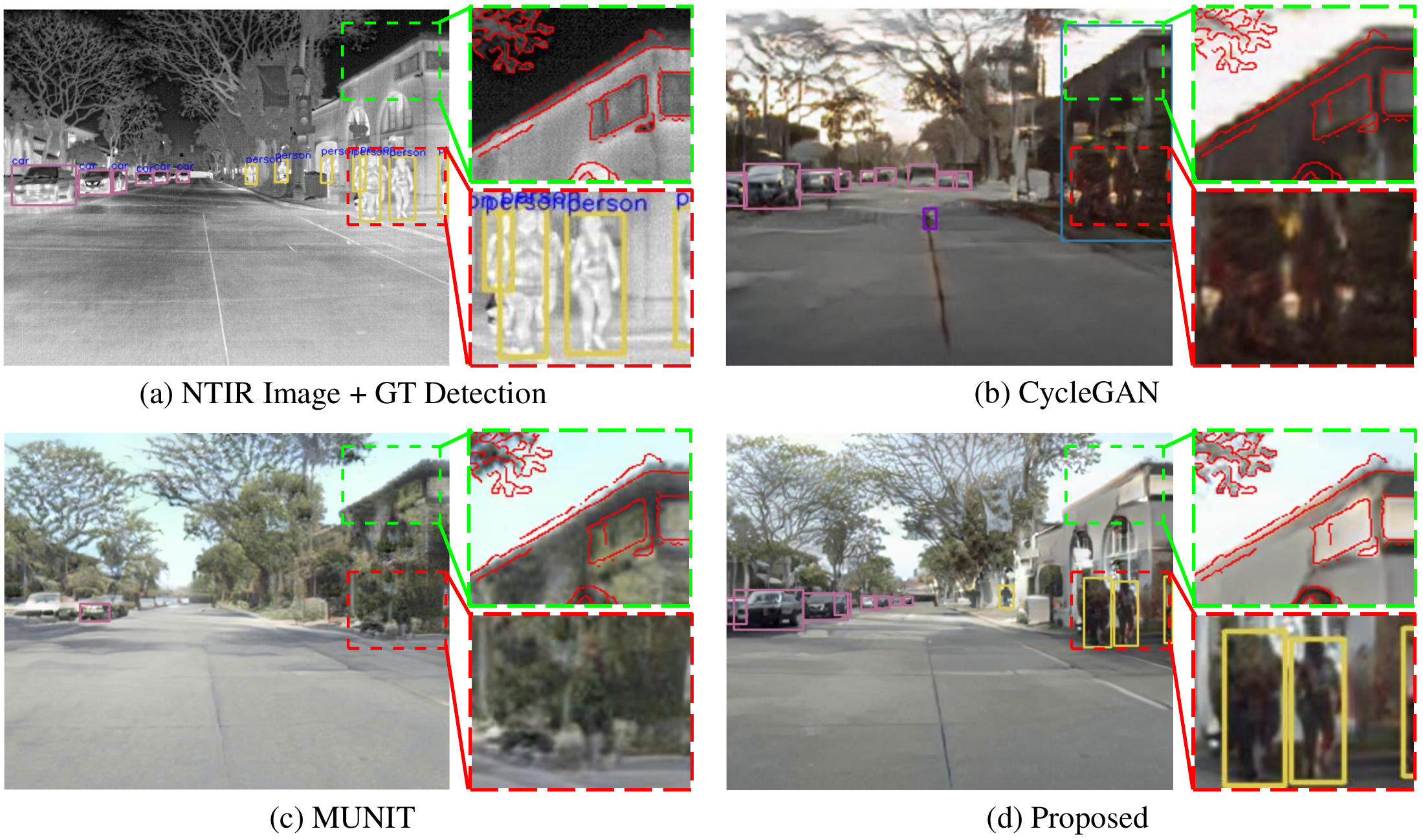}
\caption{Visual comparison of detection results and geometric consistency. In the 
second column, the green dashed box is the zoomed-in result of the fusion between the 
corresponding region and the Canny edge of the original NTIR image, and the red dashed 
box is the zoomed-in result of the detection. Please zoom in to check more details on the 
content and quality.}
\label{fig_intro}
\end{figure}

Although impressive results have been obtained through unsupervised I2I translation methods, 
the problem of geometrical distortion during translation is still under-addressed, and how to 
utilize contextual information to reduce local semantic ambiguity for NTIR image encoding is 
still under-explored. As shown in Fig. \ref{fig_intro}, both CycleGAN and MUNIT fail at generating plausible 
pedestrians and keeping the edges consistent with the original image. To address the above 
mentioned problems, we propose a new GAN model incorporated with toP-down attEntion And gRadient aLignment 
based on ToDayGAN, referred to as PearlGAN.

Visual attention allows us to interact with our environment by selectively attending to 
the information that is relevant to our behavior. Studies \cite{2007-AAAS-Saalmann} have 
shown that the spatial information about a scene 
extracted by the posterior parietal cortex (PPC) forms the basis for feedback signals to 
highlight neural responses as early along the visual pathway as the primary visual cortex. 
Such feedback modulated filtering helps reduce information overload and enables effective 
continuous visual search by directing our attention to specific locations in the 
visual field \cite{1999-BRR-Vidyasagar}. 
Inspired by this attentional feedback, we devise a top-down guided attention (TDGA) module 
with an elaborate attentional loss to reduce local semantic ambiguity in infrared image coding. 
Specifically, the TDGA module first uses average pooling at different scales to extract 
spatial contextual information of different receptive fields. Then, the maximum receptive 
field extracted features serve as coarse global information (analogous to the scene 
information extracted by PPC), which directs the attention of different receptive field 
units to specific spatial locations from coarse to fine, so as to achieve hierarchical 
scene coding. The introduced attentional loss function includes attentional diversity loss and attentional cross-domain 
conditional similarity loss, which are used to encourage a diverse distribution of attentional 
maps at the spatial level and feature level, respectively. For 
the edge distortion problem, we design a structured gradient alignment 
loss to penalize edge shift or disappearance during the image translation process.

Given that image colorization is a multi-solution problem, a natural question is how to evaluate 
the accuracy of colorization without ground truth. Indeed, 
an acceptable colorized TIR image requires not only realistic features but also a rigorous 
preservation of scene content, which is difficult to fully assess using 
the FID score \cite{2017-NIPS-Heusel}. Therefore, we use a semantic segmentation model pre-trained on real DC images 
to evaluate the semantic preservation performance of various colorization methods. In addition, we 
propose a new metric called APCE (i.e., Average Precision of Canny Edges under multi-threshold 
conditions) to evaluate geometric consistency between the original and translated images. 

To measure the semantic segmentation performance of the translated images, we perform pixel-level 
category annotation on a subset of NTIR images from the FLIR \cite{2019-FLIR-FA} and 
KAIST \cite{2015-CVPR-Hwang} datasets. Exhaustive experiments on these two datasets show that the proposed method not only 
achieves plausible colorization, but also outperforms other state-of-the-art translation methods in 
terms of image content and geometry preservation. The main contributions of this study 
are summarized as follows: 
\begin{itemize}
  \item We design a top-down guided attention module 
  and a corresponding attentional loss to achieve hierarchical attention distribution and reduce 
  local semantic ambiguity of image encoding using contextual information.
  \item We introduce a structured gradient alignment loss to reduce edge 
  distortion in the NTIR2DC task.
  \item We annotate a subset of FLIR and KAIST datasets with pixel-level category 
  labels, which may catalyze research on the colorization and semantic segmentation of 
  NTIR images.
  \item To the best of our knowledge, we are the first to propose evaluation 
  metrics to assess the semantic and edge preservation of NTIR2DC methods.
  \item Extensive experiments on the NTIR2DC task show that the proposed model significantly 
  outperforms other image translation methods in terms of semantic preservation and edge 
  consistency.
\end{itemize}

In the remainder of this paper, Section II summarizes related work about TIR image colorization and I2I 
translation. Section III introduces the architecture of the proposed PearlGAN. Section IV presents 
our experiments on FLIR and KAIST datasets. Section V draws the conclusions of our work.


\section{Related Work}

In this section, we briefly review previous work on TIR image colorization, 
unpaired image-to-image translation and feature aggregation with attention pyramid.

\subsection{TIR Image Colorization}

Recently, witnessing the success of deep learning based 
approaches in various computer vision tasks \cite{2016-He-CVPR,2020-TMM-Cao,2017-He-ICCV}, 
an increasing number of researchers have focused their efforts on the area of TIR 
image colorization, which can be categorized into 
supervised \cite{2018-Berg-CVPRW,2018-Qayynm-IBCAST,2019-Wang-ICIP,
2020-Abbott-CVPRW,2020-Almasri-Arxiv,2020-Bhat-ICCES,2020-Kuang-IPT,2020-Wang-PRL} 
and unsupervised \cite{2018-Nyberg-ECCV} methods. The supervised colorization methods 
require paired infrared and visible images, and theycolorize TIR 
images by minimizing the distance between the synthesized image and the corresponding RGB image. 
For example, Berg \textit{et al.} \cite{2018-Berg-CVPRW} used separate luminance and chrominance loss 
to constrain the mapping of infrared to visible RGB images. \cite{2020-Bhat-ICCES} 
and \cite{2020-Kuang-IPT} combined pixel-level content loss and adversarial loss to realize 
the colorization of thermal infrared images. In order to enhance edge information in 
the image translation process, Wang \textit{et al.} \cite{2019-Wang-ICIP,2020-Wang-PRL} 
jointly encoded infrared images and their Canny edge maps. 

However, due to the rapid changes in specific scenarios (e.g., traffic scenes), it is extremely 
difficult to collect TIR and visible image pairs with perfect pixel to pixel correspondence, 
which limits the practicality of supervised methods. Without requiring paired samples, 
unsupervised colorization methods usually use GAN models to minimize the 
distance between the distribution of the translated image and the real RGB image. 
For example, Nyberg \textit{et al.} \cite{2018-Nyberg-ECCV} utilized the CycleGAN \cite{2017-CVPR-Zhu} model 
to realize unpaired infrared-visible image translation. In general, despite the impressive results obtained 
with the previous methods, semantic encoding entanglement and geometric distortion in the 
NTIR2DC task is still under-addressed.

\subsection{Unpaired Image-to-Image Translation}

Unpaired image-to-image translation aims to use unpaired training samples 
to learn the mapping between two different image domains. 
Zhu \textit{et al.} \cite{2017-CVPR-Zhu} made the earliest effort 
to get rid of aligned image pairs by using cycle consistency loss, leading to the recent surge 
of interest in methods for unpaired image-to-image translation \cite{2017-NIPS-Liu,2019-ICLR-Kim,2020-CVPR-Chen}. 
Based on the shared-latent space assumption, Liu \textit{et al.} \cite{2017-NIPS-Liu} presented 
a general framework named UNIT to further extend the cycle-consistency constraint. 
Subsequently, MUNIT \cite{2018-ECCV-Huang} and 
DRIT++ \cite{2020-IJCV-Lee} were proposed to improve the diversity of synthesized images 
by learning a disentangled representation with a domain-invariant content space and a
domain-specific style space. Recent works further boosted the generation performance of night-to-day 
image translation by introducing multiple discriminators \cite{2019-ICRA-Anoosheh} 
or decoders \cite{2020-ECCV-Zheng}. Compared with the implicit edge consistency constraint 
using cycle consistency loss, the structured gradient alignment loss proposed in this work explicitly 
constrains the gradient structure of the synthesized image, to avoid the possible edge 
misalignment repair in the inverse mapping process. 

\subsection{Feature Aggregation with Attention Pyramid}

Recently, due to its superiority in multi-scale feature selection, a strategy that merges features 
with an attention pyramid has been widely used in salient object detection \cite{2019-CVPR-Wang,2019-CVPR-Zhao,2020-TIP-Zhang}, 
face recognition \cite{2020-CVPR-Wang} and video classification \cite{2019-AAAI-Xie}. 
For example, Wang \textit{et al.} \cite{2019-CVPR-Wang} proposed a pyramid attention module, which extends 
the regular attention mechanisms with multiscale information to improve saliency representation. 
Ni \textit{et al.} \cite{2020-AAAI-Ni} proposed a Double Attention Module to extract 
multi-scale attentive features, which were fused through a Pyramid Upsampling Module. To extract the 
most discriminative semantic regions for domain adaptation, Li \textit{et al.} \cite{2020-ECCV-Li} 
adopted a task-oriented guided spatial attention pyramid learning strategy to aggregate 
hierarchical semantic information in feature maps of the pyramid. Hu \textit{et al.} \cite{2020-ECCV-Hu} 
constructed a pyramid of local self-attention blocks to achieve localization of image manipulations. 

The module most similar to the proposed TDGA module might be the Cascaded Pyramid 
Attention (CPA) module \cite{2020-TIP-Zhang}, as both progressively estimate high-resolution 
attention maps from coarse to fine. However, different from the CPA module that uses only 
the largest scale attention map to select input features, the TDGA module uses attention maps of 
different scales to selectively weight the corresponding groups of feature maps. In terms of purpose, the 
TDGA module focuses on the hierarchical attention of the whole scene and the disentanglement 
of spatial features, whereas the CPA module aims to strengthen the regional features of specific objects.

\section{Proposed Method}

In this section, we first present the overview of the proposed PearlGAN. Subsequently, 
we briefly explain the ToDayGAN model \cite{2019-ICRA-Anoosheh} as our baseline and its variants for 
TIR image colorization. Then the details of the proposed TDGA module are described. 
Next, we explicate the elaborate attentional loss, including attentional diversity (AD) loss 
and attentional cross-domain conditional similarity (ACCS) loss. After that, the structured 
gradient alignment (SGA) loss to enable edge consistency during translation is explained. 
Finally, we illustrate the total loss of the proposed PearlGAN.

\begin{figure*}[!t]
\centering
\includegraphics[width=0.9\textwidth]{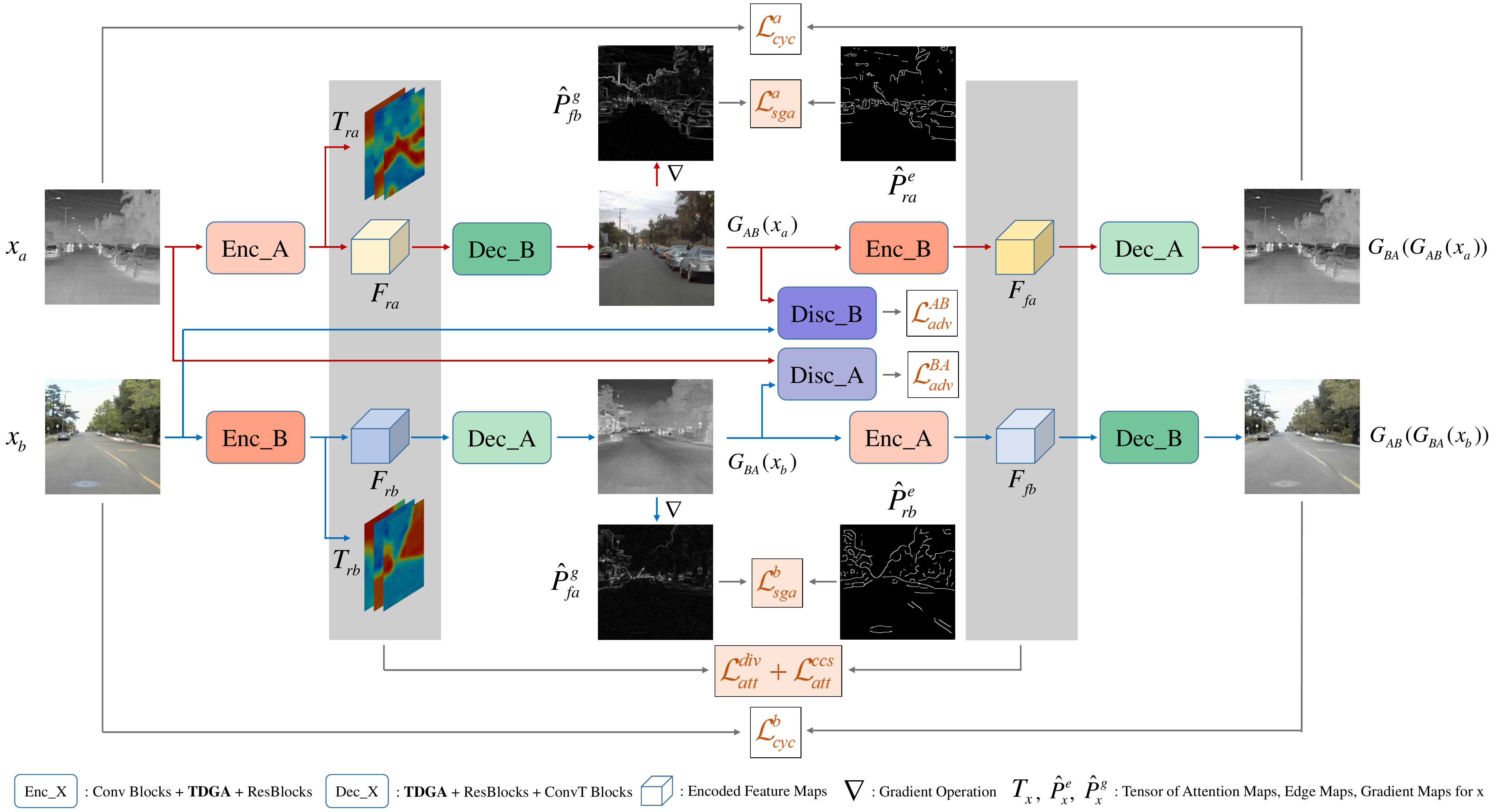}
\caption{The overall architecture of the proposed method. $x_a$ and $x_b$ respectively denote 
random images from nighttime TIR domain A and daytime visible domain B. The gray arrows 
indicate the composition of the loss function, and the red and blue arrows correspond to 
the forward computation of domain A and domain B, respectively. The abbreviations for 
Encoder, Decoder and Discriminator are Enc, Dec and Disc, respectively. We first propose a 
TDGA module and an elaborate attentional 
loss $\left (\mathcal{L} _{att}^{div} + \mathcal{L} _{att}^{ccs}\right )$ to reduce the semantic encoding 
ambiguity during translation. Then, a structured gradient alignment 
loss $\left (\mathcal{L} _{sga}^{a} + \mathcal{L} _{sga}^{b}\right )$ is introduced to encourage geometric 
consistency between the translated images and input images.}
\label{fig_model}
\end{figure*}

\subsection{Overview and Problem Formulation}

The overall framework is shown in Fig. \ref{fig_model}. We choose the ToDayGAN \cite{2019-ICRA-Anoosheh} model as the baseline 
model, which consists of a pair of generators and discriminators with a total objective 
function including adversarial loss $\mathcal{L}_{adv}$ and cycle-consistency loss $\mathcal{L}_{cyc}$. 
The generator consists of an encoder and a decoder. Due to its limitations on the NTIR2DC 
task, we first adapt the ToDayGAN model using the existing loss 
functions and modules to obtain a new model called ToDayGAN-TIR. Based on the ToDayGAN-TIR 
model, we introduce a novel TDGA module and an attentional loss to reduce the coding ambiguity 
of NTIR images, and a SGA loss $\mathcal{L}_{sga}$ to reduce the 
edge distortion of translation results. The final model obtained is called PearlGAN. 
Attentional loss consists of an AD loss $\mathcal{L}_{att}^{div}$ and 
an ACCS loss $\mathcal{L}_{att}^{ccs}$.

In the rest of the paper, the NTIR and DC image sets will correspond to 
domain A and domain B, respectively. Taking the translation from domain A to domain 
B as an example, the input image pair 
of domain A and B is denoted as $\left \{ x_{a}, x_{b} \right \} $, the generator $G_{AB}$ contains an encoder of 
domain A and a decoder of domain B, and the discriminator $D_B$ aims to distinguish 
the real data $x_b$ from the translated data $G_{AB} \left ( x_{a}  \right ) $. Similarly, the inverse 
mapping includes the generator $G_{BA}$ and the discriminator $D_A$.

\subsection{Revisiting and Improving ToDayGAN Model}
\subsubsection{Revisiting ToDayGAN Model}
ToDayGAN modifies the ComboGAN model \cite{2018-CVPRW-Anoosheh} to improve the performance of retrieval-based localization 
tasks. It has two paired generator-discriminator modules, which are capable of learning mappings 
between nighttime and daytime visible images. The generator networks are the same as the 
networks used in CycleGAN, whereas each discriminator network contains three copies to focus 
on different aspects (i.e., texture, color, and gradients) of the input. Similarly to CycleGAN, 
ToDayGAN model is supervised by two losses, i.e., adversarial loss and cycle-consistency loss $\mathcal{L}_{cyc}^{ori}$. 
Differently, it chooses Relativistic Loss \cite{2018-Arxiv-Jolicoeur} adapted to 
least-squares GAN loss as adversarial loss $\mathcal{L} _{adv}$.

\subsubsection{Improving ToDayGAN Model for TIR Image Colorization}

There are three distinct limitations when directly applying ToDayGAN to the NTIR2DC 
task. First of all, the translated results usually exhibit color dot artifacts, which degrade 
the naturalness of the image. Secondly, we observe that sometimes the reconstructed 
TIR image is different from the input TIR image, although the cycle-consistency loss value is still 
small. Furthermore, the training process is unstable. 

To avoid the aforementioned drawbacks, we adjust some designs of the ToDayGAN model. 
Referring to StyleGAN2 \cite{2020-CVPR-Karras}, droplet artifacts may be caused by the 
generator's desire to use instance normalization to achieve scale-specific controls. 
But unlike StyleGAN2, which uses the weight demodulation module to replace instance 
normalization, we design a novel alternative that removes the artifacts while bringing fewer 
changes to the model structure. The main idea is replacing the last two instance normalization 
layers of the decoder with two group normalization \cite{2018-ECCV-Wu} layers, and then 
introducing the total variation \cite{1992-PD-Rudin} loss $\mathcal{L} _{tv}$ to punish the 
noise in the translated image. 
The problem of inaccurate representation of cycle-consistency loss motivates us to think about 
the expressive ability of loss function. We notice that the temperature of some background 
categories (e.g., tree, building and road) in the night environment is low and the 
difference among them is small, resulting in low 
contrast and blurred boundaries among local areas of the TIR image. We speculate that this 
may be the reason why the cycle-consistency loss based on the L1-norm is not sensitive to the 
structural differences between TIR images. Consequently, we additionally introduce SSIM \cite{2004-TIP-Wang}
loss $\mathcal{L}_{ssim}$ to supplement the evaluation of structural differences between 
images\footnote{We use the implementation provided by \url{https://www.cnpython.com/pypi/pytorch-msssim.}}. 
Then the improved cycle-consistency loss can be expressed as:
\begin{equation}
  \label{L_adv}
  {\mathcal{L}_{cyc}} = {\lambda_{cyc}}\mathcal{L} _{cyc}^{ori} + {\lambda _{ssim}}{\mathcal{L} _{ssim}},   
\end{equation}
where $\lambda_{cyc}$ and $\lambda_{ssim}$ are loss weights. For the problem of unstable 
training, similar to \cite{2019-ICLR-Kim,2020-CVPR-Chen}, we add spectral 
normalization \cite{2018-Arxiv-Miyato} after each convolutional layer of the discriminator. Finally, 
combining the above three adjustments, we can obtain the variant model ToDayGAN-TIR, which serves as 
a starting point for the designed PearlGAN.

\subsection{Top-Down Guided Attention Module}

The ToDayGAN-TIR model still does not solve the local encoding ambiguity problem of TIR images. 
In the absence of a wide range of contextual information, understanding local areas of 
TIR images is more challenging than visible images, such as recognizing the wheel of a car. 
We speculate that this may be the reason for the confusing encoding of local features 
of TIR images by existing image translation models. Inspired by the attentional 
feedback mechanism in biological vision \cite{1999-BRR-Vidyasagar}, we design the TDGA 
module and the customized attentional loss, which not only gradually uses a wide 
range of contextual information to assist the semantic 
perception of complex regions, but also reduces the spatial entanglement of features.

\begin{figure}[!t]
\centering
\includegraphics[width=3.45in]{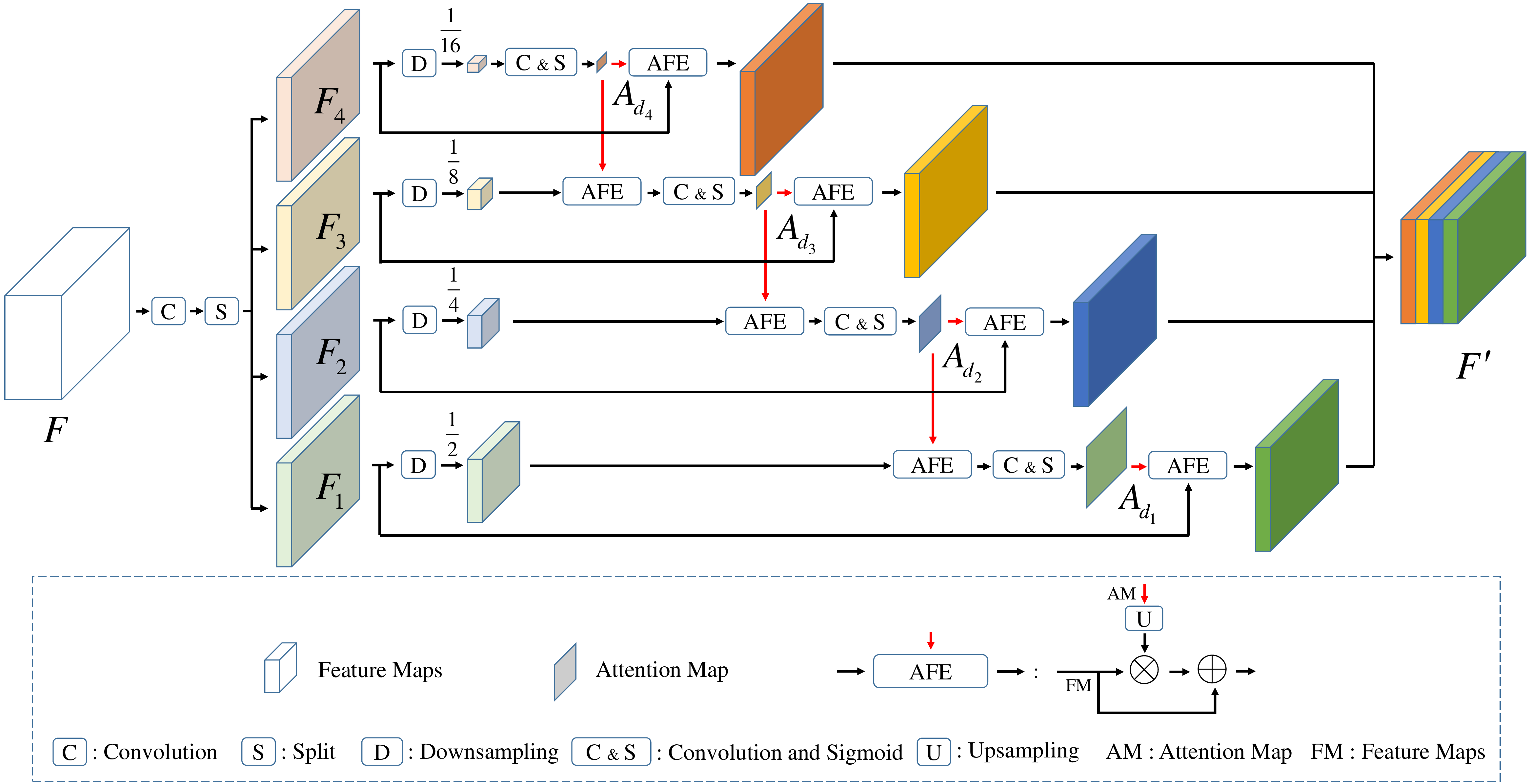}
\caption{Illustration of the proposed TDGA module. AFE indicates attention-directed feature 
enhancement.}
\label{fig_tdga}
\end{figure}

An illustration of the TDGA module is provided in Fig. \ref{fig_tdga}. For a given input feature map $F\in \mathbb{R}^{c\times h\times w}$, 
we first combine a convolutional layer to separate features of different scales to obtain 
four sets of features $\left \{ F_{1}, F_{2},F_{3},F_{4} \right \}$, and the channel number 
of each set of features is $\frac{c}{4}$. 
Then, a statistical feature pyramid $F_{d_{s} } \in \mathbb{R}^{\frac{c}{4} \times \frac{h}{2^{s} }\times \frac{w}{2^{s} }}$ 
is obtained by using four scales of stacked $2\times$ average pooling operation to represent the statistical 
features of different receptive field ranges, where $s\in \left \{ 1,2,3,4 \right \} $ indexes 
the pyramid scale. Next, the features of the large receptive field progressively predict the 
attention features of small scales, and realize the attention estimation and feature 
extraction from coarse to fine. Specifically, we apply a $3\times 3$ convolution and Sigmoid 
activation function to $F$, which produces a 2D spatial attention map $A_{F} \in \mathbb{R}^{h\times w}$:
\begin{equation}
  \label{att}
  A_{F} = Att(F) = \sigma (conv(F;\hat{\theta})),   
\end{equation}
where $\sigma(.)$ denotes the Sigmoid activation function, and $conv(.;\hat{\theta})$ denotes 
a convolutional layer with the parameter $\hat{\theta}$. In order to use a wide range of 
contextual information to gradually predict the attention map, we first utilize the smallest 
resolution feature map $F_{d_{4}}$ to infer an attention map $A_{d_{4}}$ by 
Eq. (\ref{att}). The attention cue mined at this scale is applied over the feature of the 
next scale, 
which guides the features of the next scale to pay attention to contextual signal and 
predicts the attention map more completely. Then, the attention map at the second feature 
resolution (i.e., $8\times $ down-sampling) is generated by: 
\begin{equation}
  \label{gafe}
  A_{d_{3}} =Att\left(F_{d_{3}} + F_{d_{3} }\odot \left ( A_{d_{4} }\uparrow_{2}\right)\right) ,   
\end{equation}
where $\odot$ denotes element-wise multiplication with channel-wise broadcasting, 
and $\uparrow_{2}$ means the $2\times$ spatial up-sampling operation. By analogy, we can infer the 
other attention maps $A_{d_{2}}$ and $A_{d_{1}}$ of different scales by Eq. (\ref{gafe}). Through 
the cascaded architecture, higher-resolution features can focus on the areas with more integral 
semantics under the guidance of the lower-resolution attention cues, which provide a wide 
range of contextual information. To preserve the hierarchical attention features, the 
attention cue mined at each scale is applied over the corresponding feature of the original 
spatial resolution, which are further merged in a cascading manner. The output feature maps 
are generated by:
\begin{equation}
  \label{tdga}
  \begin{array}{*{20}{c}}
    {{F{}' } = concat(({F_4} + {F_4} \odot ({A_{{d_4}}}{ \uparrow _{16}})),({F_3} + {F_3} \odot ({A_{{d_3}}}{ \uparrow _8})),} \\ 
    {({F_2} + {F_2} \odot ({A_{{d_2}}}{ \uparrow _4})),({F_1} + {F_1} \odot ({A_{{d_1}}}{ \uparrow _2})))}, 
  \end{array} 
\end{equation}
where $concat\left ( . \right ) $ represents feature channel concatenation. Since the middle layer of the network 
may need enough contextual information to realize the transition from low-level features to 
high-level semantics, we simply put the TDGA module before the residual module in the encoder and 
decoder to assist with feature disentanglement.  

\subsection{Attentional Loss}

To further ensure that the TDGA module can produce a hierarchical attention distribution, we 
introduce an elaborate attentional loss. Attentional loss is composed of AD loss and ACCS loss, 
which encourage hierarchical coding of features at the spatial distribution level and feature level, respectively. Since the global attention can 
further benefit robust perception of the whole scene with stronger generalization ability, 
the designed attentional loss is only applied to the three finer scales (i.e., $8\times ,4\times ,2\times$ down-sampling). 
In other words, the number of attention scales applied to attentional loss, denoted 
as $n_{a}$, is set to three throughout the rest of the paper. We first restore attention maps of 
different scales to the original spatial resolution (i.e., $h\times w$) by an up-sampling 
operation, and then the attention maps are further concatenated to obtain the cascaded attention 
tensor $T \in \mathbb{R} ^{n_{a} \times h\times w}$:
\begin{equation}
  \label{fuse_am}
  T=concat\left ( {A_{{d_3}}}{ \uparrow _8},{A_{{d_2}}}{ \uparrow _4},{A_{{d_1}}}{ \uparrow _2} \right ) .   
\end{equation}
The details of the introduced attentional loss are described below.

\subsubsection{Attentional Diversity Loss}

AD loss enforces the mutual exclusion and completeness of attention at multiple 
scales in space, and the concrete form is designed as follows:
\begin{equation}
  \label{att_div}
  \begin{array}{*{20}{c}}
    {\mathcal{L}_{att}^{div} = \alpha  \times \frac{1}{{h \times w}} \times \sum\limits_{i,j} {[(1 - \max_{k\in S }{T_{k,i,j}} )} } \\ 
    { + \beta  \times {{(\sum\limits_k {{T_{k,i,j}}}  - 1)}^2}]},
  \end{array}   
\end{equation}
where $\alpha$ and $\beta$ are tuning coefficients to enforce the range of loss being 0 
to 1, $S= \left \{ 1,...,n_{a}\right \}$ 
represents the set for the scales of attention maps, and ${T_{k,i,j}}$ denotes the 
attention weight located at $\left ( i,j \right )$ for 
the $k_{th}$ channel of the attention tensor. In the square brackets, the two parts before and 
after the plus sign encourage the maximum value and sum value of the multi-scale attention maps 
at the same spatial position to be as close to one as possible. In this way, the attention of different 
scales will focus on different spatial regions, which benefits feature disentanglement. 
To normalize the output range to $\left( 0,1 \right)$, the coefficients $\alpha$ and $\beta$ are 
set to $\frac{1}{2}$ and $\frac{1}{4}$, respectively.

\subsubsection{Attentional Cross-Domain Conditional Similarity Loss}

Although AD loss encourages the diversified spatial distribution of attention, a lack of 
semantic level constraints might lead to irregular distribution of attention and hinder 
hierarchical coding of features. Hence, we introduce ACCS loss not only to encourage 
the compactness of the attentional features, but also to encourage the conditional similarity 
of the same-scale attentional features across the domains, which will be 
beneficial to generate a synthesized image with richer details and more natural texture. We 
first define the encoding features (i.e., the output of the encoder) of real NTIR, real DC, 
fake NTIR and fake DC images as $F_{ra}$, $F_{rb}$, $F_{fa}$ and $F_{fb}$, respectively, and all of 
them have the same dimensions as F. Similar to the cascaded attention tensor T, the 
attention tensors of real NTIR and real DC images can be denoted as $T_{ra}$ and $T_{rb}$ through 
Eq. (\ref{fuse_am}), and the dimensions of the two are the same as T. We define the attention 
feature as the weighted sum of the attention map and the coding feature in the spatial 
dimension and normalize it with its L2 norm. For example, we suppose the attention feature of a fake 
DC image is denoted as $V_{fbra} \in \mathbb{R}^{n_{a}\times c} $, which is calculated by 
using $F_{fb}$ and $T_{ra}$, and its $k_{th}$ scale 
component $V_{fbra}^k\in \mathbb{R}^{1\times c}$ is defined as:
\begin{equation}
  \label{att_fea_norm}
  V_{fbra}^k = \frac{{\widetilde{V} _{fbra}^k}}{{{{\left\| {\widetilde{V}_{fbra}^k} \right\|}_2}}},   
\end{equation}
where $\left \| . \right \| _{2}$ represents the L2 norm, and the unnormalized 
feature $\widetilde{V} _{fbra}^k$ can be formulated as:
\begin{equation}
  \label{att_fea}
  \widetilde{V} _{fbra}^k = \frac{{GAP({F_{fb}} \odot T_{ra}^k)}}{{GAP(T_{ra}^k)}},   
\end{equation}
where $GAP(.)$ denotes the globally averaged pooling operation, and $T_{ra}^k$ represents 
the $k_{th}$ channel of attention tensor $T_{ra}$. Then we can obtain different scale 
components of the attention feature (i.e., $V_{rara}$, $V_{rbrb}$, $V_{farb}$ and $V_{fbra}$) 
for real--fake image pairs 
by utilizing Eq. (\ref{att_fea_norm}). Next, we can calculate the similarity between attention 
features with cosine distance. For example, it is assumed that the similarity of attention 
features in the visible spectral domain can be defined 
as $Q_{b} \in \mathbb{R}^{n_{a}\times n_{a}}$, which is given by:
\begin{equation}
  \label{fea_sim}
  {Q_b} = Mm({V_{rbrb}},{({V_{fbra}})^T}) = {V_{rbrb}} \otimes {({V_{fbra}})^T},   
\end{equation}
where $\otimes$ is the matrix multiplication, and $\left( . \right )^{T}$ represents the 
transpose of the matrix. Based on the observation of the regularity of the 
spatial distribution of semantic categories in traffic scene images, for NTIR-DC image pairs 
with similar semantic distribution, we expect that the similarity of their attention features 
at the same scales should be greater than that of cross scales. Let $diag\left ( Q_b\right)\in \mathbb{R}^{n_{a}\times 1}$ 
be the diagonal elements of the similarity matrix $Q_b$, where the $k_{th}$ element represents 
the same-scale similarity of the $k_{th}$ scale, and let $M\left ( Q_{b}  \right ) $ denote the 
row-wise maximum of similarity matrix $Q_b$. Then we formulate ACCS loss function in the visible 
spectral domain as: 
\begin{equation}
  \label{accs_b}
  \begin{array}{*{20}{c}}
    {\mathcal{L}_{accs}^b = \frac{{{W_Q} \otimes [M({Q_b}) - diag({Q_b})]}}{{\sum\limits_{k \in S} {{{({W_Q})}_k}} }}} \\ 
    {{\kern 1pt} {\kern 1pt}{\kern 1pt}{\kern 1pt}{\kern 1pt}{\kern 1pt}{\kern 1pt}{\kern 1pt}{\kern 1pt}{\kern 1pt}{\kern 1pt}{\kern 1pt}{\kern 1pt}{\kern 1pt}{\kern 1pt}{\kern 1pt}{\kern 1pt}{\kern 1pt}{\kern 1pt}{\kern 1pt}{\kern 1pt}{\kern 1pt}{\kern 1pt}{\kern 1pt}{\kern 1pt}{\kern 1pt}{\kern 1pt}{\kern 1pt}{\kern 1pt}{\kern 1pt}{\kern 1pt}{\kern 1pt}{\kern 1pt}{\kern 1pt}{\kern 1pt}{\kern 1pt}{\kern 1pt}{\kern 1pt}{\kern 1pt}{\kern 1pt}{\kern 1pt}{\kern 1pt} {\kern 1pt} {\kern 1pt} {\kern 1pt} {\kern 1pt} {\kern 1pt} {\kern 1pt} {\kern 1pt} {\kern 1pt} {\kern 1pt} {\kern 1pt} {\kern 1pt} {\kern 1pt} {\kern 1pt} {\kern 1pt} {\kern 1pt} {\kern 1pt} {\kern 1pt} {\kern 1pt} {\kern 1pt} {\kern 1pt} {\kern 1pt} {\kern 1pt} {\kern 1pt}  + Dis({V_{rbrb}}) + Dis({V_{fbra}})}, 
  \end{array}
\end{equation}
where $W_Q\in \mathbb{R}^{1\times n_{a}}$ is an indicator vector representing the confidence of 
different scales of cross-domain attention, and $Dis\left ( . \right )$ is a distance function 
to encourage a compact distribution of attention features. Concretely, the $k_{th}$ element of $W_Q$ is 
given by:
\begin{equation}
  \label{weight_q}
  {({W_Q})_k} = \min (\max (T_{ra}^k),\max (T_{rb}^k)).
\end{equation}
The distance function $Dis\left ( . \right )$ expects the feature distance 
between scales to be as large as possible to ensure a compact distribution of features within 
the scale. For example, let $Q_{rbrb} \in \mathbb{R}^{n_{a}\times n_{a}}$ denote the matrix product between $V_{rbrb}$ and its 
transpose. Then the distance function for attention feature $V_{rbrb}$ can be formulated as:
\begin{equation}
  \label{dis}
  Dis({V_{rbrb}}) = \max (\frac{{\sum\limits_{i \ne j,i,j \in S} {{{({Q_{rbrb}})}_{ij}}} }}{{{n_a}({n_a} - 1)}},0),
\end{equation}
where the numerator represents the sum of the elements on the non-main diagonal, the 
denominator represents the number of the non-main diagonal elements, and the the purpose of non-linear 
function $\max (x,0)$ is to avoid negative values. Similarly, the ACCS loss 
in the TIR spectral domain $\mathcal{L}_{accs}^a$ can be obtained through Eq. (\ref{accs_b}). Finally, 
the complete ACCS loss function is defined as:
\begin{equation}
  \label{accs_all}
  \mathcal{L} _{att}^{ccs} = \mathcal{L} _{accs}^a + \mathcal{L} _{accs}^b,
\end{equation}

\subsection{Structured Gradient Alignment Loss}

While the TDGA module and attentional loss can improve the stability of the model for TIR 
image encoding, the problem of edge distortion in the image translation process is still 
under-resolved. We observe that edge regions in TIR images usually appear at the junction of
two regions with temperature difference, and the gradient at its corresponding position in 
the visible image is likely to be larger than the average of their neighboring gradients. 
In light of that, we propose a SGA loss to encourage the ratio of the gradient of the 
resulting image at the edge position to the maximum value of its neighborhood to be greater 
than a given threshold, while ignoring the constraints on non-edge regions.

Different from the previous methods \cite{2019-Wang-ICIP,2020-Wang-PRL}, 
the proposed SGA loss not only avoids the need for a pre-trained edge detection network, 
but also explicitly restricts the gradient structure of the translated image. Specifically, 
taking the TIR spectral domain as an example, we first use the offline edge detection method 
MCI \cite{2014-TIP-Yang} to predict the edge map of the TIR image in the training set. Then, 
we let the size of the image block be ${l_p} \times {l_p}$ and use the method of adaptive average 
pooling to randomly select an image patch $P_{ra}^e$ with edge pixels in the TIR edge map. 
The corresponding image patch in the gradient map of the fake DC image is named $P_{fb}^g$. 
Both $P_{ra}^e$ and $P_{fb}^g$ are normalized to the range $[0,1]$ by dividing the maximum 
of the corresponding image patch. Next, the SGA loss of the TIR domain can be defined as:
\begin{equation}
  \label{sga_a}
  \mathcal{L} _{sga}^a = \frac{{\sum\limits_{i = 1}^{{l_p}} {\sum\limits_{j = 1}^{{l_p}} {\max ({{(\eta  \times P_{ra}^e - P_{fb}^g)}_{ij}},0)} } }}{{\sum\limits_{i = 1}^{{l_p}} {\sum\limits_{j = 1}^{l_p^{}} {P_{ra}^e} } }},
\end{equation}
where the parameter $\eta$ is a threshold that controls edge sharpness. Due to the 
variability of the brightness range of TIR images from different datasets, we set the value of 
parameter $\eta$ to be related to the maximum intensity value of all 
infrared images in the dataset, which is denoted as $I_{max}$ and given by:
\begin{equation}
  \label{th}
  \eta  = 0.8 \times \frac{{{I_{\max }}}}{{255}},
\end{equation}
where 0.8 is an empirical threshold representing the minimum gradient ratio to get a clear 
edge after normalization, and 255 represents the maximum value of an 8-bit image. By analogy, 
we can obtain the SGA loss of the visible image domain $\mathcal{L} _{sga}^b$ by Eq. (\ref{sga_a}). 
Consequently, the SGA loss for two domains can be formulated as:
\begin{equation}
  \label{sga}
  {\mathcal{L} _{sga}} = \mathcal{L} _{sga}^a + \mathcal{L} _{sga}^b,
\end{equation}
Therefore, the proposed SGA loss can punish the result of inconspicuous or disappearing edges 
to maintain the consistency of edges between two domains.

\subsection{Objective Function}

In summary, the full objective function of PearlGAN can be written as:
\begin{equation}
  \label{total_loss}
  \begin{array}{*{20}{c}}
    {{\mathcal{L}_{all}} = {\mathcal{L}_{adv}} + ({\lambda _{cyc}}\mathcal{L}_{cyc}^{ori} + {\lambda _{ssim}}{\mathcal{L}_{ssim}}) + {\lambda _{tv}}{\mathcal{L}_{tv}}} \\ 
    { + {\lambda _{att}}(\mathcal{L}_{att}^{div} + \mathcal{L}_{att}^{ccs}) + {\lambda _{sga}}{\mathcal{L}_{sga}},} 
  \end{array}
\end{equation}
where $\lambda_{cyc}$, $\lambda_{ssim}$, $\lambda_{tv}$, $\lambda_{att}$ and $\lambda_{sga}$ are 
loss weights. Referring to CycleGAN \cite{2017-CVPR-Zhu}, $\lambda_{cyc}$ is set to 10, and we 
empirically set the value of $\lambda_{tv}$ to be half of $\lambda_{cyc}$ to reduce color dot 
artifacts. Without loss of generality, the values of $\lambda_{ssim}$ and $\lambda_{att}$ are 
both set to one. However, the naturalness of the translated image is sensitive to the 
parameter $\lambda_{sga}$. If $\lambda_{sga}$ were to be set to one, the generated image 
would have sharp 
edges but the naturalness would be heavily degraded. Hence, we simply set the value of $\lambda_{sga}$ 
to 0.5 without more trials.

\section{Experiments}
In this section, we first introduce the datasets and evaluation metrics for the NTIR2DC task. 
Subsequently, we explain the annotation of the test images and the implementation details of the 
model. Then, experimental results on FLIR and KAIST datasets are presented. Next, we 
perform an ablation analysis to verify the validity of the proposed module and loss functions. 
At last, some discussion of the experimental results is presented.

\subsection{Datasets and Evaluation Metrics}
\subsubsection{Datasets}
Experiments are conducted on the FLIR and KAIST datasets to demonstrate the effectiveness of our 
PearlGAN. The FLIR Thermal Starter Dataset \cite{2019-FLIR-FA} provides an annotated TIR 
image set and non-annotated RGB image set for training and validating object detection 
models. According to the lighting conditions of RGB images, we finally obtain 5447 DC images 
and 2899 NTIR images for training, and 490 NTIR images in the validation set for model 
evaluation.

\begin{figure}[!t]
\centering
\includegraphics[width=2.5in]{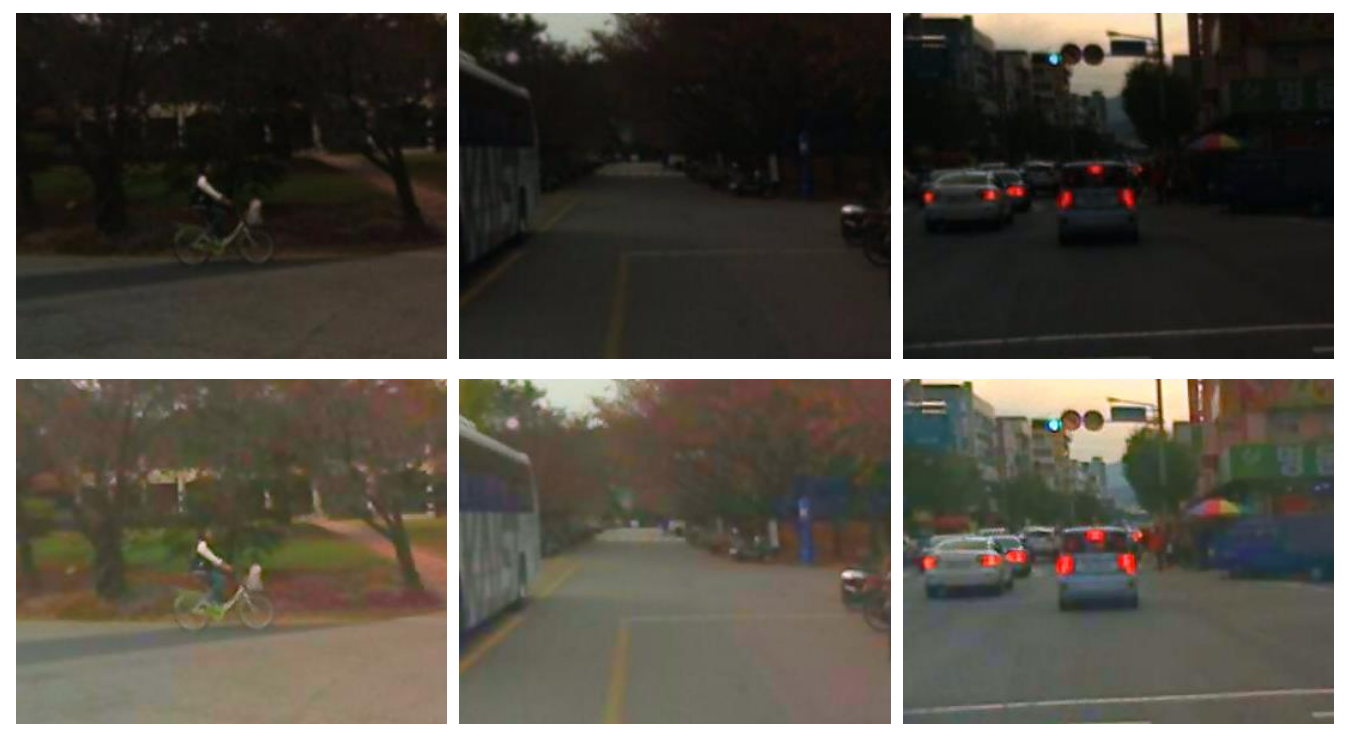}
\caption{KAIST image enhancement results achieved by SRLLIE \cite{2018-TIP-Li}. The first 
and second rows are the original image and the enhanced image, respectively.}
\label{fig_enh}
\end{figure}

The KAIST Multispectral Pedestrian Detection Benchmark \cite{2015-CVPR-Hwang} provides 
somewhat aligned color and thermal image pairs captured in day and night. Due to the low 
brightness of some DC images, as shown in Fig. \ref{fig_enh}, we enhance all the training DC images in 
KAIST using the SRLLIE \cite{2018-TIP-Li} method to improve the image quality. We finally 
obtain 1674 DC images and 1359 NTIR images as training set samples, and 500 NTIR images in 
the test set for evaluating semantic and edge 
consistency\footnote{See \url{https://github.com/FuyaLuo/PearlGAN/} for specific sample 
selection.}. The sample size of the test set used for pedestrian detection experiments 
is 611. 

Due to differences in sensor imaging resolution and processing of image registration, black 
filled areas appear on both sides of some images in the FLIR and KAIST datasets. In order to 
remove irrelevant regions, we first resize all training set images to $500\times 400$ 
resolution, and then perform center crop to get $360\times 288$ resolution input images.

\subsubsection{Evaluation Metrics}
An ideal NTIR image colorization method should preserve the image content at all 
levels, i.e., from scene level layout to fine-grained edges. Therefore, inspired by \cite{2020-ECCV-Zheng}, 
we perform three vision tasks to evaluate the coloring performance of NTIR images, 
including semantic segmentation, object detection, and edge detection.

Intersection-over-Union (IoU) \cite{2015-IJCV-Everingham}, which is the ratio of the 
intersection of the predicted segmentation map with the ground truth to their union, is a 
widely used metric for semantic segmentation. We take the mean IoU (mIoU) over all 
categories to indicate the overall performance of the model. 

For object detection, average precision (AP) \cite{2010-IJCV-Everingham} is defined as the average detection precision 
under different recalls. In the case of multiple categories, the mean 
AP (mAP) averaged over all object categories is typically used as the final metric of performance. 

Since the model may generate more details to obtain plausible DC images, it is sub-optimal 
to use traditional metrics (e.g., F-score, which considers both precision and recall) to evaluate 
the edge consistency of the NTIR2DC task. Therefore, we propose a new metric 
called APCE to evaluate the preservation of edges in the source 
NTIR image. APCE is the average precision of Canny edges under different threshold conditions. 
Given the $j_{th}$ high threshold $\mu_{j}$, the Canny edge maps of 
the $i_{th}$ test image and its corresponding output image are denoted as $X_{i}\left(\mu _{j}\right)$ 
and $Y_{i}\left(\mu_{j}\right)$, respectively. Then, we define the APCE of the entire test set 
as:
\begin{equation}
  \label{apce}
  APCE=\frac{1}{n_{i} } \frac{1}{n_{j} } \sum_{i=1}^{n_{i} } \sum_{j=1}^{n_{j} } \frac{X_{i}\left(\mu _{j}\right)\cap Y_{i}\left(\mu _{j}\right)}{X_{i}\left(\mu _{j}\right)},
\end{equation}
where $n_{i}$ and $n_{j}$ represent the total number of images and thresholds, respectively. 

\subsection{Image Annotation and Implementation Details}
\subsubsection{Image Annotation}

\begin{figure}[!t]
\centering
\includegraphics[width=3.45in]{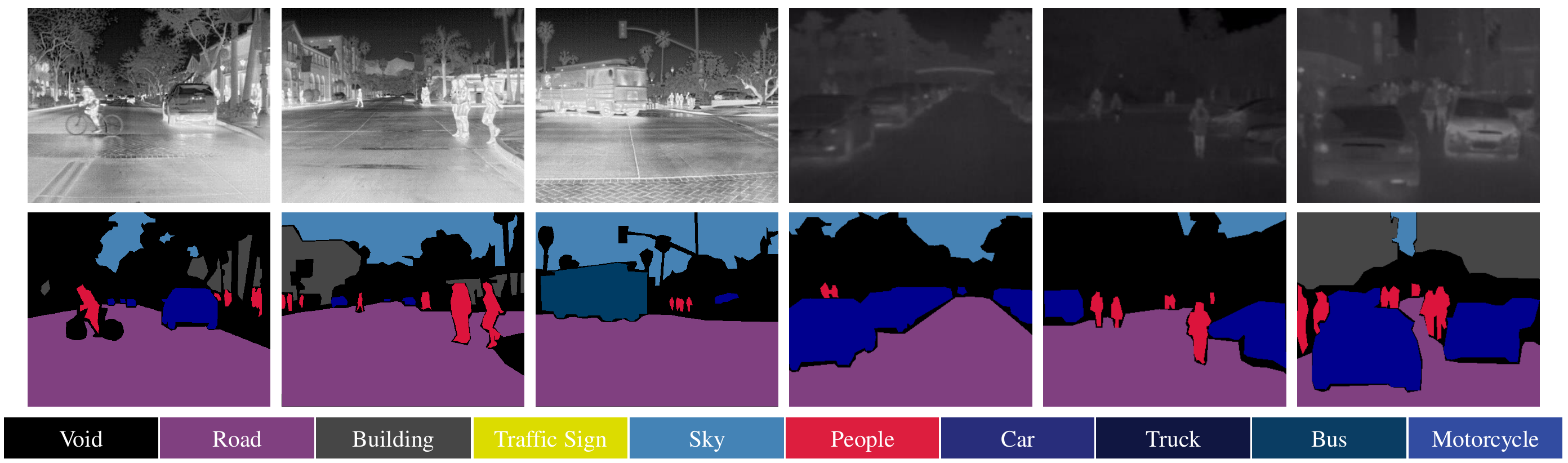}
\caption{Examples of NTIR images from the FLIR (the first three columns) and KAIST 
(the last three columns) datasets and our annotated segmentation masks.}
\label{fig_annotate}
\end{figure}

To evaluate the semantic preservation performance of the NTIR2DC task, a subset of 
FLIR and KAIST datasets are selected and annotated with their 
pixel-level category labels. Due to the low contrast and ambiguous boundaries 
of NTIR images, as shown in Fig. \ref{fig_annotate}, we define nine categories and use the $LabelMe$ 
toolbox\footnote{\url{https://github.com/CSAILVision/LabelMeAnnotationTool.}} to annotate only 
their identified corresponding regions. The labeled categories are road, building, traffic 
sign, sky, people, car, truck, bus and motorcycle.

\begin{table*}[htbp]
  \centering
  \caption{Semantic Segmentation Results of the Synthesized Images Obtained by Different 
  Translation Methods on FLIR Dataset. All Numbers are in $\%$}
    \begin{tabular}{ccccccccccc}
      \toprule
          & Road  & Building & Traffic sign & Sky   & Person & Car   & Truck & Bus   & Motorcycle & mIoU \\ \hline
    Reference NVC images & 95.5  & 62.8  & \textbf{6.4} & 63.4  & 40.7  & 51.4  & 0.0   & \textbf{0.4} & 0.0   & 35.6  \\
    CycleGAN \cite{2017-CVPR-Zhu} & 95.6  & 39.1  & 0.0   & 90.8  & 60.8  & 78.0  & 0.0   & 0.0   & 0.0   & 40.5  \\
    UNIT \cite{2017-NIPS-Liu}  & 96.2  & 60.3  & 0.2   & 92.1  & 64.5  & 71.5  & 0.0   & 0.0   & \textbf{14.6} & 44.4  \\
    MUNIT \cite{2018-ECCV-Huang} & 96.0  & 27.5  & 0.0   & 92.8  & 49.6  & 64.3  & 0.0   & 0.0   & 0.0   & 36.7  \\
    ToDayGAN \cite{2019-ICRA-Anoosheh} & 95.7  & 47.2  & 0.0   & 85.3  & 56.8  & 75.4  & 0.0   & 0.0   & 0.0   & 40.0  \\
    UGATIT \cite{2019-ICLR-Kim} & 94.3  & 18.4  & 0.0   & 89.0  & 23.7  & 59.0  & 0.0   & 0.0   & 0.0   & 31.6  \\
    DRIT++ \cite{2020-IJCV-Lee} & 97.3  & 29.4  & 0.0   & 78.4  & 28.0  & 78.9  & 0.0   & 0.0   & 0.0   & 34.7  \\
    ForkGAN \cite{2020-ECCV-Zheng} & 94.4  & 55.1  & 0.0   & 90.7  & 60.5  & 76.1  & 0.0   & 0.0   & 0.0   & 41.9  \\
    Proposed & \textbf{97.7} & \textbf{73.1} & 0.0   & \textbf{93.4} & \textbf{73.2} & \textbf{82.7} & \textbf{0.1} & 0.0   & 0.0   & \textbf{46.7} \\
    \bottomrule
    \end{tabular}%
  \label{tab_flir_seg}%
\end{table*}%

\subsubsection{Experiment Settings and Implementation Details}

Since there is little available source code for the NTIR2DC task, we compare the proposed 
PearlGAN with other state-of-the-art I2I translation methods, including CycleGAN \cite{2017-CVPR-Zhu}, 
UNIT \cite{2017-NIPS-Liu}, MUNIT \cite{2018-ECCV-Huang}, DRIT++ \cite{2020-IJCV-Lee} and  
UGATIT \cite{2019-ICLR-Kim}, and low-light enhancement 
methods such as ToDayGAN \cite{2019-ICRA-Anoosheh} and ForkGAN \cite{2020-ECCV-Zheng}. We 
follow the instructions of these methods in order to make a fair setting for comparison.

The proposed PearlGAN is implemented using PyTorch. We train the models using the Adam 
optimizer \cite{2014-Arxiv-Kingma} with $\left ( \beta _{1} ,\beta _{2}  \right ) =\left ( 0.5,0.999 \right ) $ on 
NVIDIA RTX 3090 GPUs. The batch size is set to one for all experiments. The learning rate begins 
at 0.0002, is constant for the first half of training and decreases linearly to zero during 
the second half of training. The total number of training epoches for the FLIR and KAIST datasets are 80 and 120, 
respectively. Due to the need to focus on optimizing the cycle-consistency loss early in the 
training, the SSIM loss and ACCS loss are set to start working after about 50K 
iterations (i.e., around the 10th and 30th epochs of the FLIR and KAIST dataset training, 
respectively) in order to stabilize the training process. In Eq. (\ref{sga_a}), the side 
length $l_{p}$ of the image patch is set to 32, and the gradient ratio thresholds $\eta$ for the FLIR 
and KAIST datasets are obtained as 0.8 and 0.44, respectively, from Eq. (\ref{th}). For data 
augmentation, we flip the images horizontally with a probability of 0.5, and randomly 
crop them to $256\times 256$.

Due to the lack of pixel-level semantic labels in the FLIR and KAIST datasets, we perform domain adaptive 
semantic segmentation using the Cityscape dataset \cite{2016-CVPR-Cordts} and the advanced 
scene adaptation method \cite{2019-IJCAI-Zheng} on real DC images from both datasets, and the 
final trained models are used for evaluation of the semantic segmentation performance 
of the translated images. To measure the generalizability of the features of the objects in 
the synthesized images, we utilize the YOLOv4 \cite{2020-Arxiv-Bochkovskiy} model pre-trained 
on the MS COCO dataset \cite{2014-ECCV-Lin} as an object detection evaluation method. For 
the APCE evaluation metric, the high threshold of Canny edge detection\footnote{We use the 
implementation provided by \url{https://www.mathworks.com/help/images/ref/edge.html?lang=en.}} ranges 
from 0.01 to 0.99 with an interval of 0.01, and the low threshold is taken as half of the 
high threshold.

\subsection{Experiments on FLIR Dataset}
For a comprehensive evaluation of our method, we conduct three experiments on FLIR datasets, 
including semantic segmentation, object detection and edge preservation.

\subsubsection{Semantic Segmentation}

\begin{figure}[!t]
\centering
\includegraphics[width=3.45in]{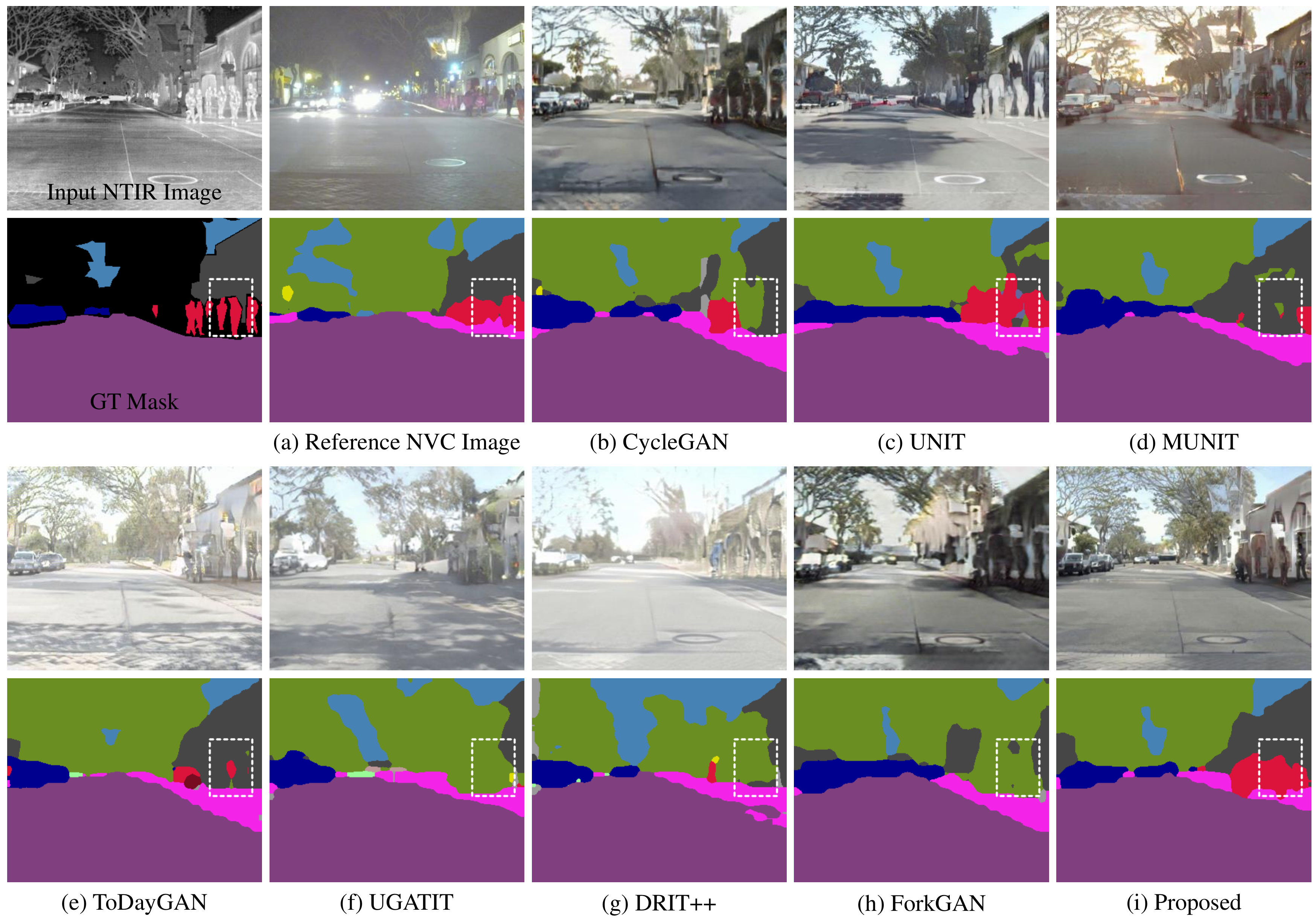}
\caption{The visual translation (the first row) and segmentation performance (the second row) 
comparison of different methods on the FLIR dataset. Please zoom in to check more details on the 
content and quality. The area covered by the white dotted box is worth attention.}
\label{fig_seg_flir}
\end{figure}

The translated results and corresponding segmentation outputs from using various methods on 
the FLIR dataset are presented in Fig. \ref{fig_seg_flir}. Column (a) lists the reference 
nighttime visible color (NVC) image and its semantic segmentation output. Although NVC images 
can provide somewhat reasonable layout cues, small cars in the distance with strong lighting 
are extremely challenging for the segmentation model to identify. In contrast, the translated 
images obtained by most translation methods can facilitate the localization of distant small 
cars. However, as shown by the white dashed box in the segmentation 
masks, CycleGAN, MUNIT, UGATIT, DRIT++ and ForkGAN fail to generate plausible pedestrians, 
and the translated images of UNIT and TDG do not guarantee complete pedestrian mask 
prediction. In contrast, our method has a stronger ability to preserve scene layout 
and generate plausible textures (e.g., buildings and cars). Table \ref{tab_flir_seg} reports 
a quantitative comparison of the semantic consistency performance on the 
FLIR dataset. The proposed method achieves the highest mIoU ($46.7\%$) among 
all the methods. In addition, our approach achieves the best results 
in the translation of semantic regions in five categories: road, building, sky, person, 
and car. Note that due to the small number of samples containing trucks, buses and motorcycles in the 
training set, as well as the small area occupied by traffic signs, all methods have poor 
translation results for these four categories.

\subsubsection{Object Detection}

\begin{figure}[!t]
\centering
\includegraphics[width=3.45in]{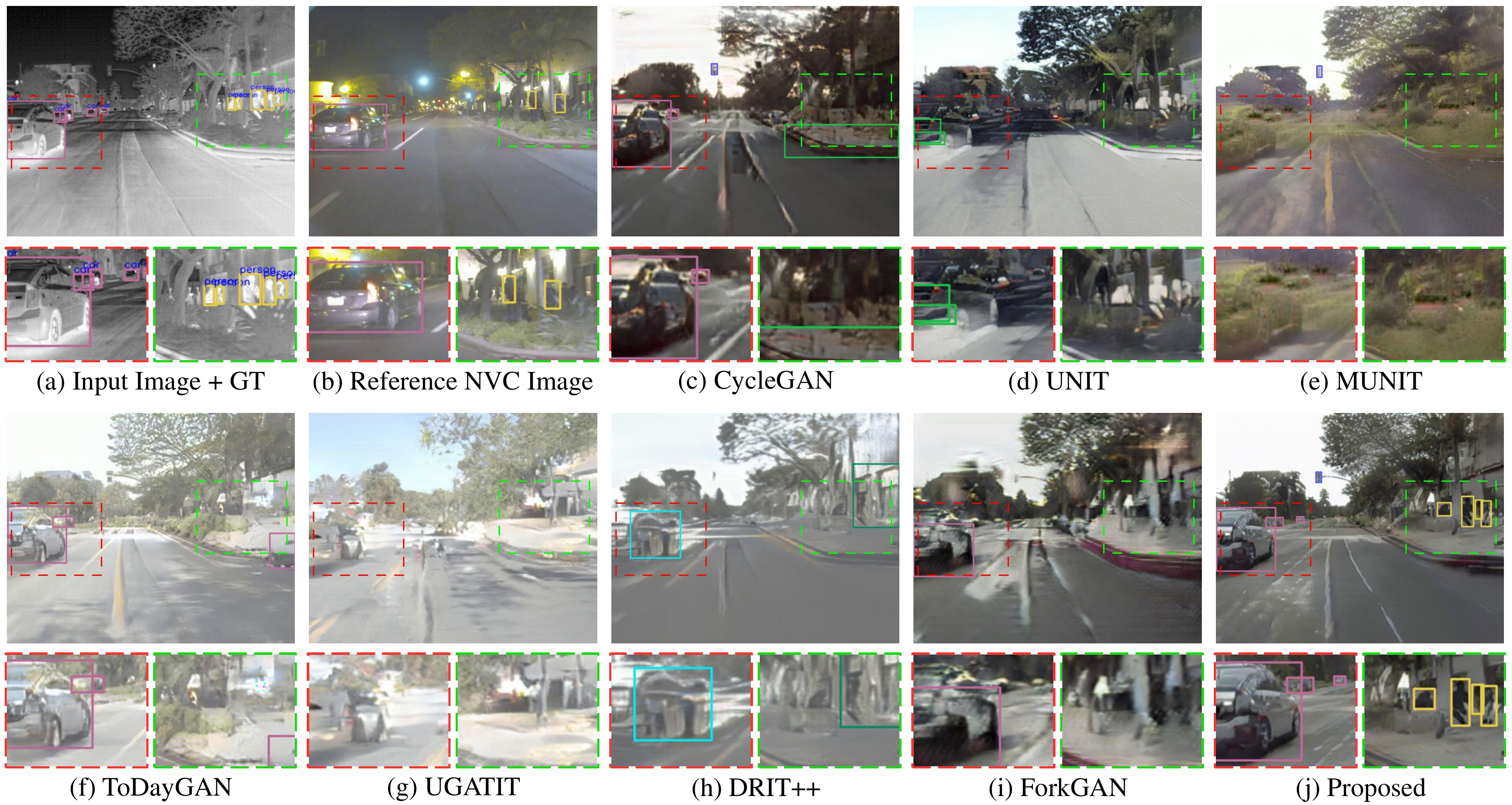}
\caption{Visual comparison of detection results on the FLIR dataset by YOLOv4 model \cite{2020-Arxiv-Bochkovskiy}. The parts covered by red and 
green dashed boxes show the enlarged cropped region in the corresponding image. Colors in the 
detection results that do not intersect with GT represent undefined categories of FLIR 
dataset identified by the detector.}
\label{fig_det_flir}
\end{figure}

Fig. \ref{fig_det_flir} shows the qualitative translation and detection result 
comparisons, wherein the second row is the zoomed in image of the corresponding area. As 
shown in the red dashed box, UNIT, MUNIT, DRIT and UGATIT cannot generate plausible cars, 
while CycleGAN, ToDayGAN and ForkGAN fail to maintain distant cars. In contrast, our method not 
only generates quite realistic cars, but also significantly outperforms other methods in terms of small object 
preservation. Furthermore, as shown in the green dashed box, all the I2I image translation 
methods are unable to generate reasonable pedestrian features to convince the general object 
detector except ours. In addition, our translation results can help more objects to be 
detected compared with the original NVC images. Since only three of the annotation 
categories (i.e., person, bicycle and car) provided by the FLIR dataset exist in the 
validation set, a quantitative comparison of the detection performance on the FLIR dataset is 
listed in Table \ref{tab_flir_det}. As shown, the detection performance on the translated images obtained 
by the proposed method substantially outperforms other methods, and our mAP result is 
almost twice as good as the second ranked method (i.e., 50.8 vs 24.8), which illustrates 
the superiority of our method in terms of object preservation.

\begin{table}[htbp]
  \centering
  \caption{Object Detection Results of the Synthesized Images Obtained by Different 
  Translation Methods on FLIR Dataset, Computed at a Single IoU of 0.50. All Numbers are in $\%$}
    \begin{tabular}{ccccc} \toprule
          & Person & Bicycle & Car   & mAP \\ \hline
    Reference NVC images & 9.8   & 2.6   & 11.5  & 8.0  \\
    CycleGAN \cite{2017-CVPR-Zhu} & 17.8  & 1.9   & 37.2  & 19.0  \\
    UNIT \cite{2017-NIPS-Liu}  & 16.3  & 9.5   & 18.3  & 14.7  \\
    MUNIT \cite{2018-ECCV-Huang} & 29.5  & 2.3   & 42.5  & 24.8  \\
    ToDayGAN \cite{2019-ICRA-Anoosheh} & 19.0  & 1.5   & 53.3  & 24.6  \\
    UGATIT \cite{2019-ICLR-Kim} & 1.9   & 0.0   & 14.6  & 5.5  \\
    DRIT++ \cite{2020-IJCV-Lee} & 16.5  & 2.2   & 46.0  & 21.6  \\
    ForkGAN \cite{2020-ECCV-Zheng} & 25.9  & 2.3   & 32.5  & 20.2  \\
    Proposed & \textbf{54.0} & \textbf{23.0} & \textbf{75.5} & \textbf{50.8} \\
    \bottomrule
    \end{tabular}%
  \label{tab_flir_det}%
\end{table}%

\subsubsection{Edge Preservation}

\begin{figure}[!t]
\centering
\includegraphics[width=3.45in]{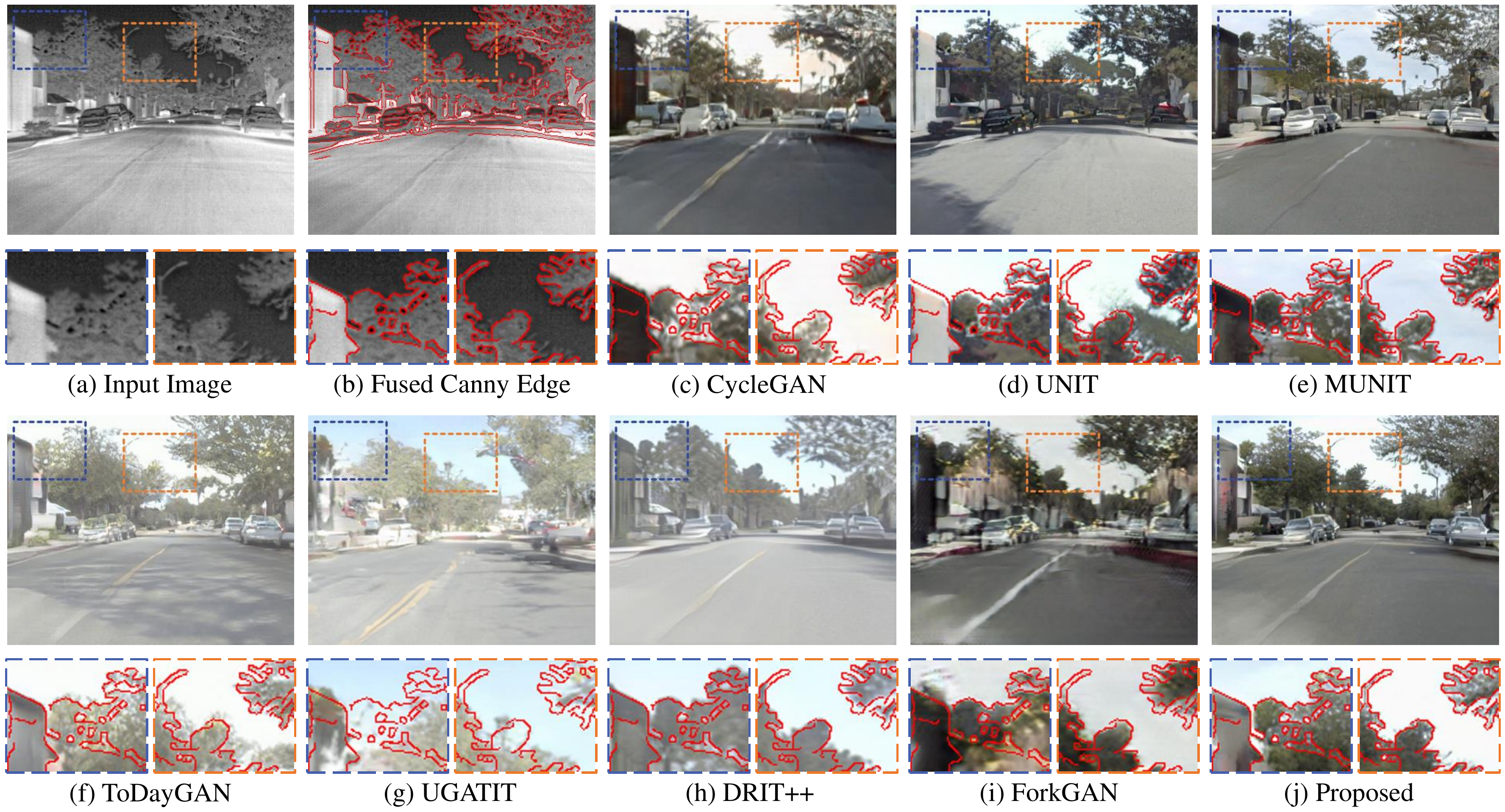}
\caption{Visual comparison of geometric consistency results on FLIR dataset. The second row 
shows the enlarged results of the corresponding regions after fusion with the edges of the 
input image. Column (b) is the result of fusing the input image with its Canny edges.}
\label{fig_edge_flir}
\end{figure}

\begin{figure}[!t]
\centering
\includegraphics[width=3.45in]{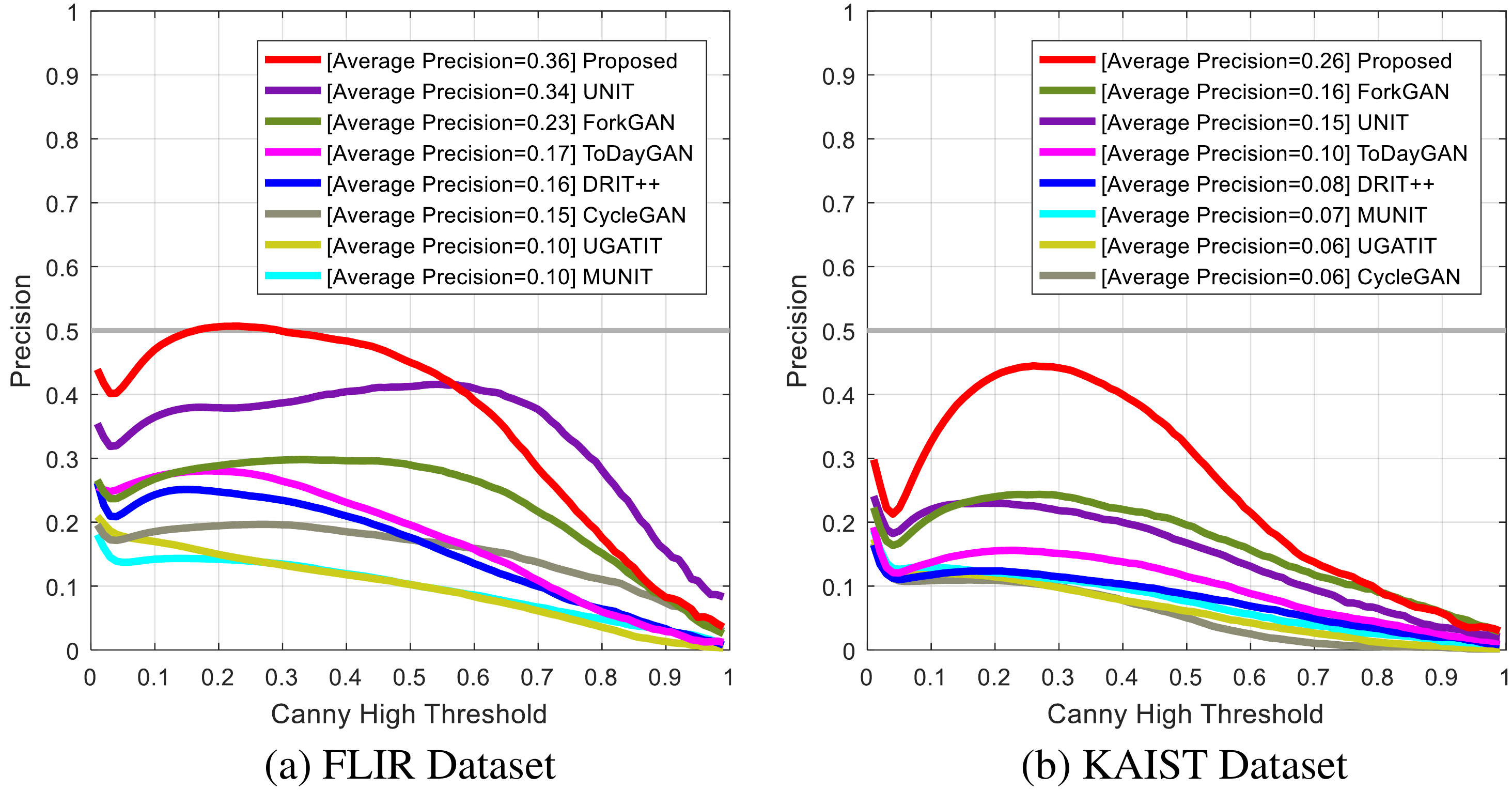}
\caption{APCE results of different translation methods on FLIR and KAIST datasets.}
\label{fig_apce}
\end{figure}

In Fig. \ref{fig_edge_flir}, we qualitatively compare the edge preservation of 
different image translation methods, and the second row shows the zoomed-in patches of 
the corresponding area fused with the edges of the original NTIR image. All compared 
I2I translation methods commonly have edges that expand outward (e.g., the 
yellow dotted box area in columns d and f) or edges that contract inward (e.g., the blue 
dotted box area in columns d, e and g). In contrast, the translated images obtained by 
the proposed method better preserve the edge structure of the original images, as shown 
in the blue dashed box in column j. For the yellow dashed box region, although a portion 
of the branches generated by our method is beyond the edges of the original 
image (i.e., the red line), these branches are present in the original NTIR image but 
are not captured by the Canny edge detector with the current parameters. 
Therefore, we investigate the edge preservation performance under different Canny 
thresholds, and the results of each method are shown in Fig. \ref{fig_apce}(a). Considering all thresholds, our method consistently keeps higher 
performance in terms of edge consistency during translation. 

\subsection{Experiments on KAIST Dataset}

To further verify the effectiveness and robustness of our proposed method, we conduct 
experiments on the challenging KAIST dataset. Due to the low resolution of early 
imaging equipment, both NTIR and DC images in the KAIST dataset are blurred and 
low contrast, which makes the NTIR2DC task even more difficult on this dataset.

\subsubsection{Semantic Segmentation}

\begin{figure}[!t]
\centering
\includegraphics[width=3.45in]{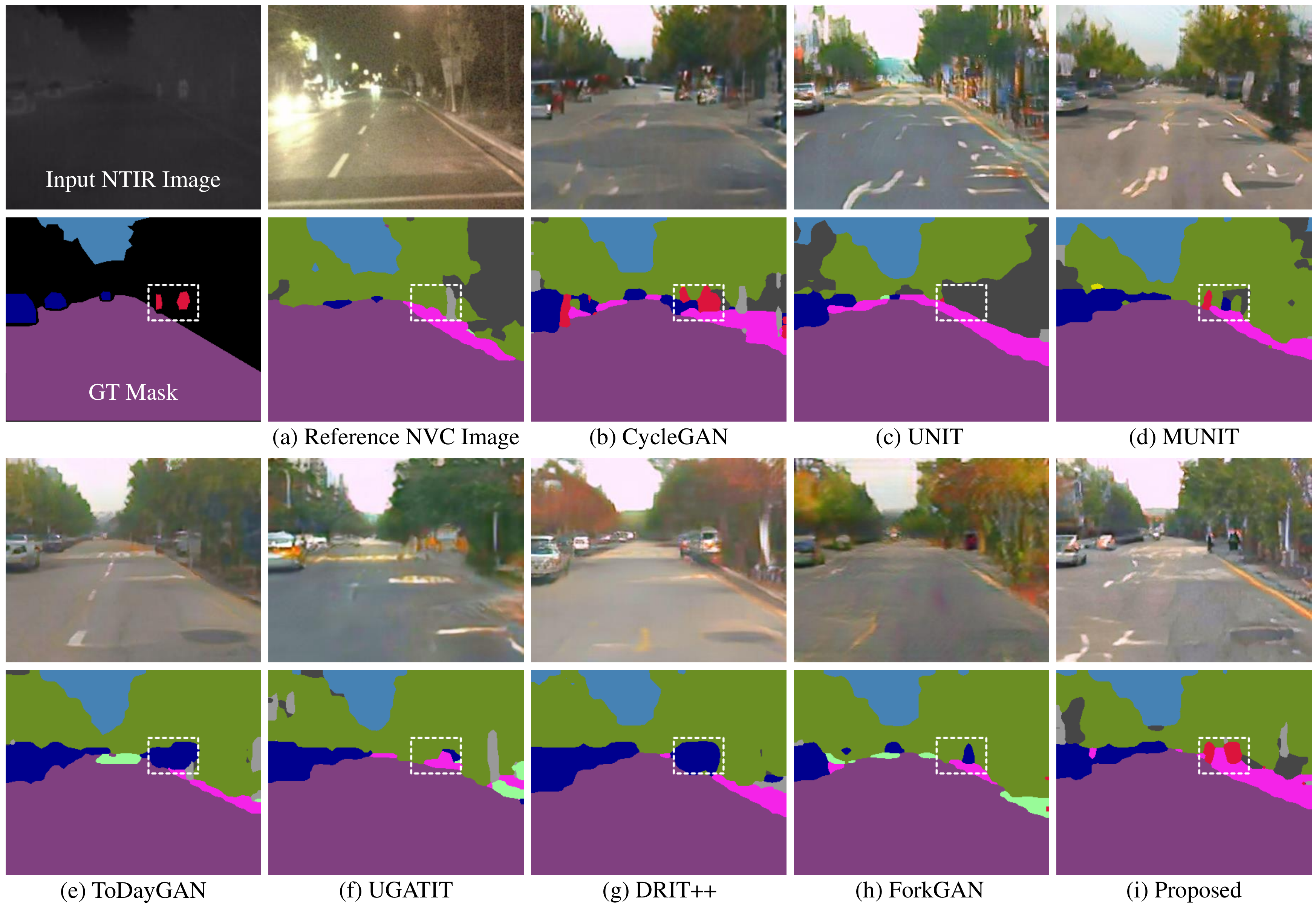}
\caption{The visual translation (the first row) and segmentation performance (the second row) 
comparison of different methods on KAIST dataset. The area covered by the white dotted box is 
worth attention.}
\label{fig_seg_kaist}
\end{figure}

Fig. \ref{fig_seg_kaist} presents the translated results and the segmentation outputs of various 
methods on the KAIST dataset. As shown in the white dashed box in the second row, the 
segmentation model is unable to recognize the pedestrians on the roadside due to the 
low illumination in the NVC image. UNIT, ToDayGAN, UGATIT, DRIT++ and ForkGAN fail 
to capture the characteristics of pedestrians. Although CycleGAN and MUNIT can make 
the segmentation model perceive pedestrians, the generation of pedestrian regions is 
incomplete (e.g., MUNIT) or overfilled (e.g., CycleGAN). However, benefitting from the 
proposed TDGA module and SGA loss, our PearlGAN can better avoid the corruption of 
objects. The quantitative comparison is reported in Table \ref{tab_kaist_seg}. After the 
alignment process with the NTIR images, the mIoU is only $33.8\%$ when we 
directly perform the semantic segmentation on the real NVC images. Note that the 
proposed PearlGAN achieves the highest mIoU among all the methods, and a significant 
improvement (i.e., + $9.3\%$) compared with that on the real NVC 
images, which indicates that PearlGAN can facilitate nighttime scene perception while 
better preserving the scene layout. Similar to the experiments on the FLIR dataset, the 
poor performance in the translation of traffic signs and vehicles other than cars is due to 
the small number of samples available for learning.

\begin{table*}[htbp]
  \centering
  \caption{Semantic Segmentation Results of the Synthesized Images Obtained by Different Translation Methods on KAIST Dataset. All Numbers are in $\%$}
    \begin{tabular}{ccccccccccc} \toprule
          & Road  & Building & Traffic sign & Sky   & Person & Car   & Truck & Bus   & Motorcycle & mIoU \\ \hline
    Reference NVC images & 92.2  & 71.8  & 0.0   & 66.3  & 15.7  & 57.7  & 0.0   & 0.2   & 0.0   & 33.8  \\
    CycleGAN \cite{2017-CVPR-Zhu} & 88.0  & 48.7  & 0.0   & 78.6  & 15.0  & 49.2  & 0.0   & 0.0   & 0.0   & 31.1  \\
    UNIT \cite{2017-NIPS-Liu}  & 94.1  & \textbf{73.9} & 3.1   & 86.6  & 36.0  & 67.7  & 0.0   & 0.0   & 1.6   & 40.3  \\
    MUNIT \cite{2018-ECCV-Huang} & 88.7  & 34.5  & 0.2   & 81.0  & 7.8   & 46.2  & 0.0   & 0.0   & 0.6   & 28.8  \\
    ToDayGAN \cite{2019-ICRA-Anoosheh} & 93.3  & 63.2  & 2.3   & 87.7  & 20.4  & 58.3  & 0.0   & 0.0   & 0.0   & 36.1  \\
    UGATIT \cite{2019-ICLR-Kim} & 90.0  & 52.2  & 1.3   & 73.3  & 16.7  & 53.0  & 0.0   & 0.0   & 0.0   & 31.8  \\
    DRIT++ \cite{2020-IJCV-Lee} & 91.2  & 71.5  & 0.0   & 73.8  & 5.1   & 56.2  & 0.0   & 0.0   & 0.0   & 33.1  \\
    ForkGAN \cite{2020-ECCV-Zheng} & 93.9  & 54.3  & 0.9   & 87.0  & 22.7  & 66.2  & 0.0   & 0.0   & \textbf{2.9} & 36.4  \\
    Proposed & \textbf{94.7} & 72.2  & \textbf{5.8} & \textbf{91.2} & \textbf{43.0} & \textbf{67.7} & 0.0   & \textbf{13.0} & 0.0   & \textbf{43.1} \\
    \bottomrule
    \end{tabular}%
  \label{tab_kaist_seg}%
\end{table*}%

\subsubsection{Pedestrian Detection}

\begin{figure}[!t]
\centering
\includegraphics[width=3.45in]{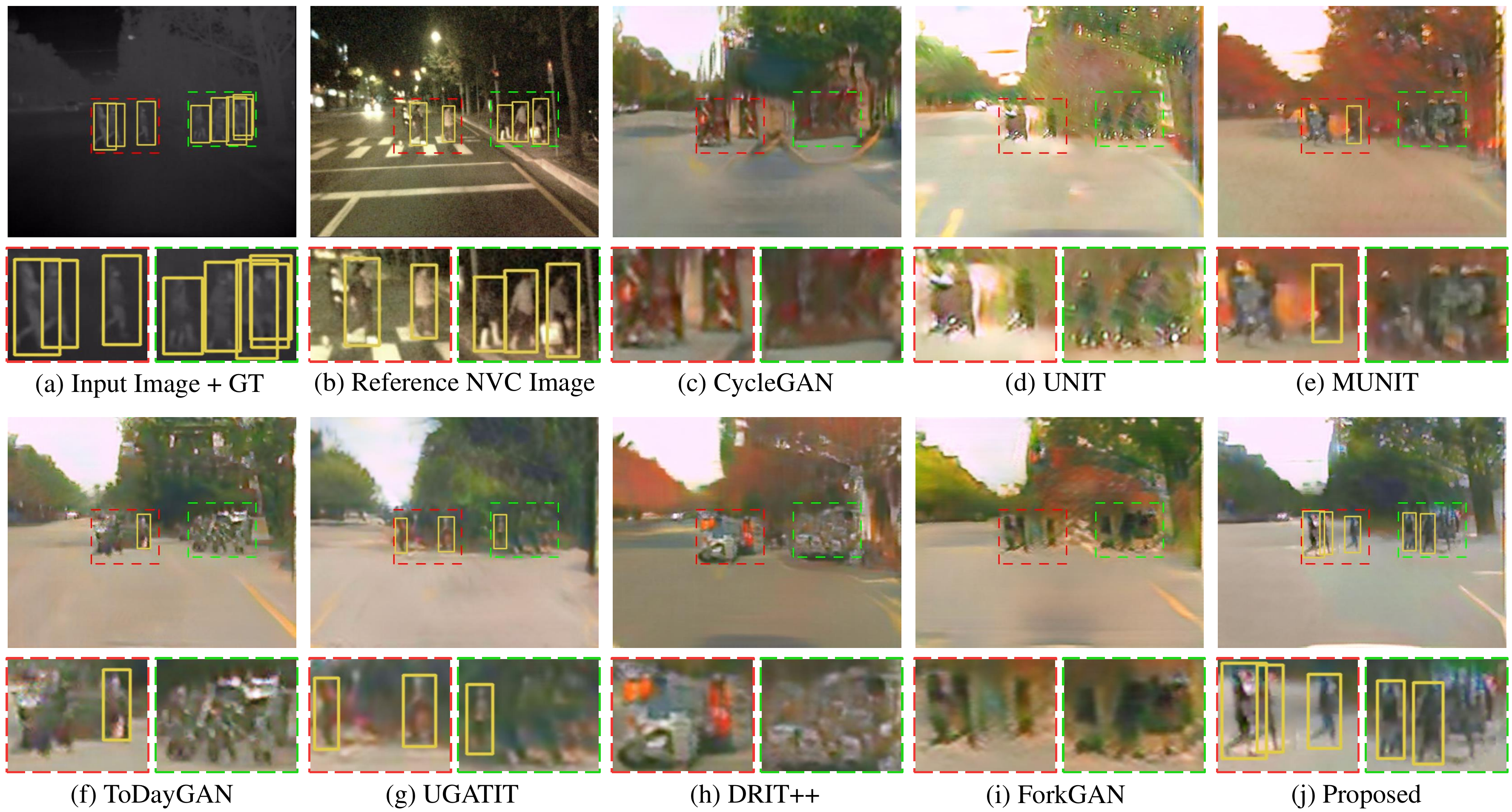}
\caption{Visual comparison of pedestrian detection results on KAIST dataset by YOLOv4 model \cite{2020-Arxiv-Bochkovskiy}. The parts covered 
by red and green dashed boxes show the enlarged cropped regions in the corresponding image.}
\label{fig_det_kaist}
\end{figure}

Since the KAIST dataset only provides bounding box annotation for pedestrians, we investigate 
the pedestrian generation performance of different I2I translation methods using YOLOv4. 
In Fig. \ref{fig_det_kaist}, we qualitatively compare the different methods. Among all the compared 
methods, only MUNIT, ToDayGAN and UGATIT generate pedestrians accepted by the detector 
despite the lack of realistic structure, whereas the other methods fail completely. In contrast, 
as shown in the red dashed box in the figure, the proposed PearlGAN can better maintain the 
pedestrian structure and perform a more reasonable translation. The quantitative comparison 
is listed in Table \ref{tab_kaist_det}. Due to the low quality of NTIR images and the small number of 
pedestrian samples in real DC images, the pedestrian detection performance on DC images 
synthesized by all translation methods is far inferior to that on the original NVC images. 
Nevertheless, the proposed method significantly outperforms the other compared translation 
methods in pedestrian transformation, and the mAP for pedestrian detection is twice as high 
as that of the second ranked method (i.e., 25.8 vs. 11.0).

\begin{table}[htbp]
  \centering
  \caption{Pedestrian Detection Results of the Synthesized Images Obtained by Different 
  Translation Methods on KAIST Dataset, Computed at a Single IoU of 0.50. All Numbers are 
  in $\%$. Top Three Results Are Marked in \textcolor[rgb]{1, 0, 0}{Red}, \textcolor[rgb]{0, 0, 1}{Blue}, 
  and \textcolor[rgb]{0, 1, 0}{Green}.}
    \begin{tabular}{cccc} \toprule
          & Precision & Recall & mAP \\ \hline
    Reference NVC images & 36.8 & 50.1 & \textcolor[rgb]{1, 0, 0}{44.2} \\
    CycleGAN \cite{2017-CVPR-Zhu} & 4.7   & 2.8   & 1.1  \\
    UNIT \cite{2017-NIPS-Liu}  & 26.7  & 14.5  & \textcolor[rgb]{0, 1, 0}{11.0}  \\
    MUNIT \cite{2018-ECCV-Huang} & 2.1   & 1.6   & 0.3  \\
    ToDayGAN \cite{2019-ICRA-Anoosheh} & 11.4  & 14.9  & 5.0  \\
    UGATIT \cite{2019-ICLR-Kim} & 13.3  & 7.6   & 3.2  \\
    DRIT++ \cite{2020-IJCV-Lee} & 7.9   & 4.1   & 1.2  \\
    ForkGAN \cite{2020-ECCV-Zheng} & 33.9  & 4.6   & 4.9  \\
    Proposed & 21.0  & 39.8  & \textcolor[rgb]{0, 0, 1}{25.8}  \\
    \bottomrule
    \end{tabular}%
  \label{tab_kaist_det}%
\end{table}%

\subsubsection{Edge Preservation}

\begin{figure}[!t]
\centering
\includegraphics[width=3.45in]{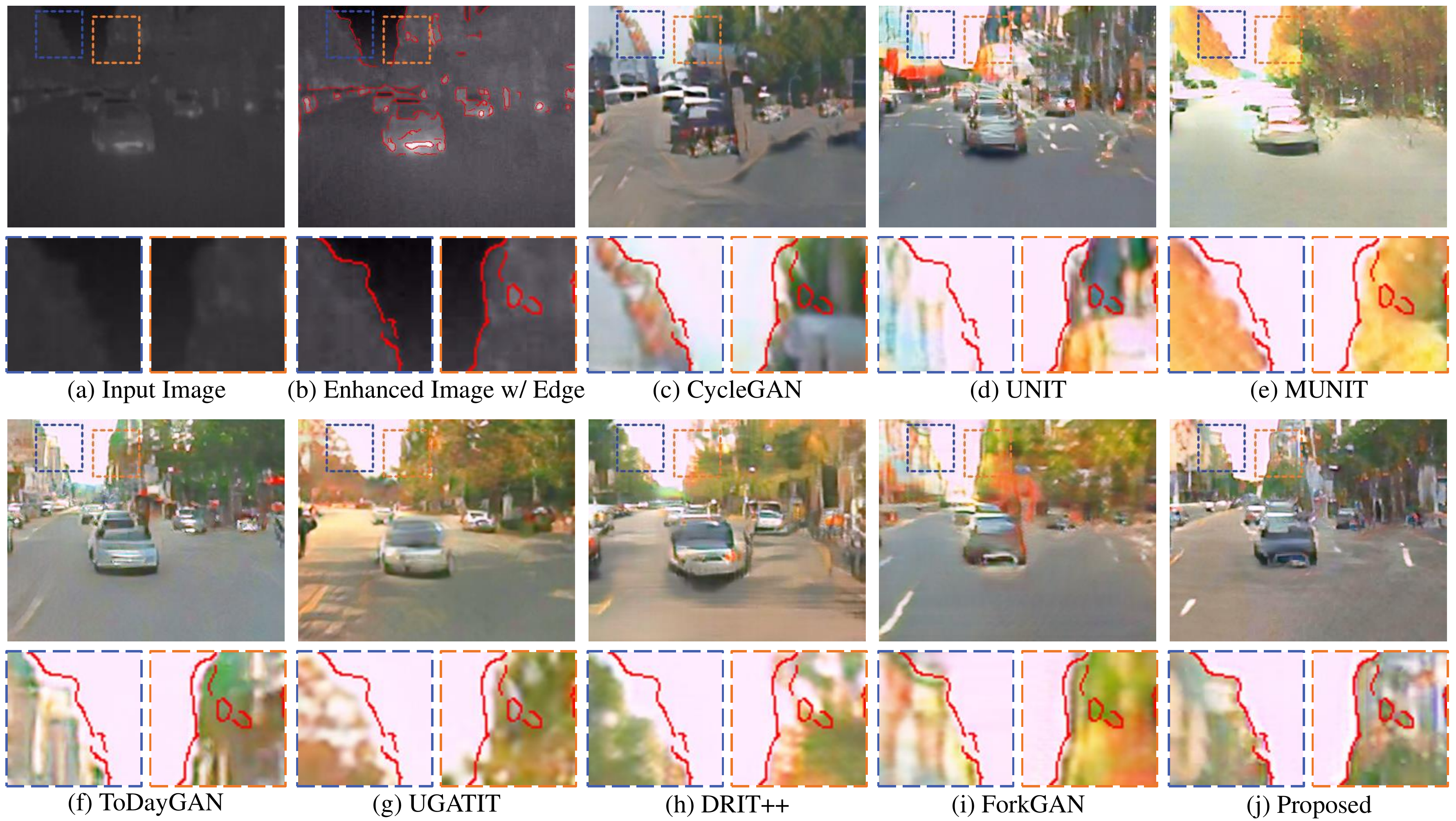}
\caption{Visual comparison of geometric consistency results on KAIST dataset. The second row 
shows the enlarged results of the corresponding regions after fusion with the edges of the 
input image. Column (b) is the result of blending the enhanced image of the input image with its 
Canny edges for better viewing.}
\label{fig_edge_kaist}
\end{figure}

The qualitative comparison of edge preservation on the KAIST dataset is shown in Fig. \ref{fig_edge_kaist}. 
For better visualization, column (b) shows the fusion results of the Canny edges of the 
original image and its enhanced image. We can observe from the figure that the compared 
translation methods all have edge shift problems, especially UNIT, MUNIT and 
DRIT++ (e.g., blue dashed box area). While the synthesized image obtained by the proposed 
method can better preserve the edge structure of the original image, as shown in column (j), 
the edges of the original NTIR image fit perfectly with the edges of the translated DC 
image. Considering different thresholds of Canny edges, the quantitative comparison of the 
edge consistency of various translation methods on the KAIST dataset is shown in 
Fig. \ref{fig_apce}(b). As shown, the edge consistency performance of the proposed method significantly surpasses 
other methods at almost all thresholds, which further illustrates the superiority of PearlGAN in edge structure preservation.

\subsection{Ablation Study}

\begin{figure*}[!t]
\centering
\includegraphics[width=1\textwidth]{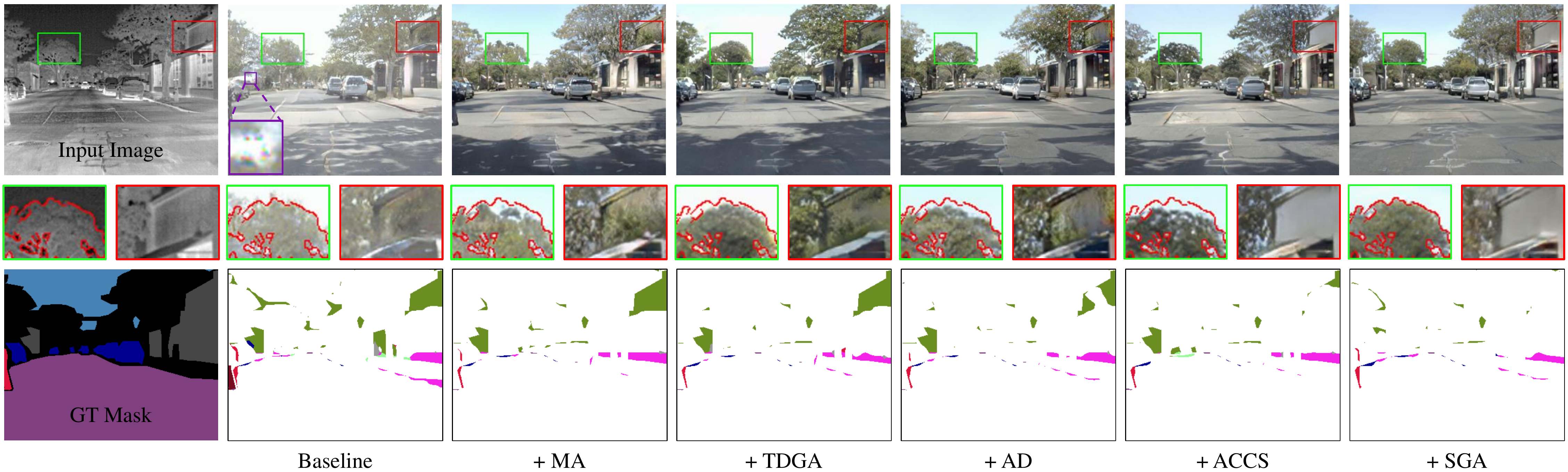}
\caption{Visual results of ablation study on FLIR dataset. The first row shows the 
translation results for different models. In the second row, the parts covered by green 
dashed boxes show the enlarged results of the corresponding regions after fusion with the 
edges of the input image, and the parts covered by red dashed boxes show the enlarged 
cropped regions in the corresponding image. The third row shows the error maps of the 
semantic segmentation results, where the white areas indicate the correct regions or 
unlabeled regions.}
\label{fig_ablation}
\end{figure*}

An ablation study is conducted on the FLIR dataset to evaluate the contribution of 
each component in PearlGAN. The results of the ablation study are presented in Table \ref{tab_aa}, 
and an example of qualitative comparison of each proposed component is 
shown in Fig. \ref{fig_ablation}. We can find that the brightness of the synthesized image obtained from 
the baseline model is so high that it produces a hazy visual effect, and there is a degree 
of content distortion in the results. In addition, the output shows artifacts of 
multiple colored dots aggregated together (i.e., the purple box in the first row). Then, as 
shown in the third column of the figure, with the help of the existing 
modules, our redesigned ToDayGAN-TIR model not only eliminates artifacts but also improves 
image quality as well as reduces content distortion. The results in the table further 
illustrate the validity of the model adaptation. In the case of 
simply introducing the TDGA module without other losses, although the model does not gain 
much in terms of content preservation, there is a significant improvement in edge 
consistency due to the coarse-to-fine feature encoding, which may help the model to 
better capture the edge information in NTIR images. In the next experiment, we study the 
influence of the attentional diversity loss. We find that the model gains a 
slight improvement in semantic preservation, when compared with the model using only the 
TDGA module. Then we investigate the effectiveness of the attentional cross-domain condition 
similarity loss. As shown in the red box in the sixth column of the figure, the 
translated image obtains a more realistic building area compared with the previous 
model. This illustrates that the introduced ACCS loss can reduce the feature 
entanglement in the model encoding process and achieve more reasonable translation. Finally, 
we explore the improvement from the proposed SGA loss. As shown 
in the figure, the edge structure of the translated image matches well with the 
original image after the SGA loss is introduced, which indicates that the proposed 
loss can effectively reduce geometric distortion. In addition, we find from the 
table that the reliable edge structure is beneficial for semantic preservation, which 
further demonstrates the superiority of the proposed architecture.

\begin{table}[htbp]
  \centering
  \caption{Quantitative Comparisons for Ablation Studies on FLIR Dataset. "MA" Means the 
  Model Adaptation to Obtain ToDayGAN-TIR.}
  \renewcommand\tabcolsep{2.5pt}
    \begin{tabular}{ccccccccc} \toprule
    Baseline & MA    & TDGA  & AD   & ACCS & SGA  & mIoU (\%) & mAP (\%) & APCE \\ \hline
    \checkmark     &       &       &       &       &       & 40.0  & 24.6  & 0.17 \\
    \checkmark     & \checkmark     &       &       &       &       & 43.9  & 45.0  & 0.27 \\
    \checkmark     & \checkmark     & \checkmark     &       &       &       & 44.3  & 46.2  & 0.34 \\
    \checkmark     & \checkmark     & \checkmark     & \checkmark     &       &       & 44.8  & 48.1  & 0.34 \\
    \checkmark     & \checkmark     & \checkmark     & \checkmark     & \checkmark     &       & 45.2  & 48.7  & 0.34 \\
    \checkmark     & \checkmark     & \checkmark     & \checkmark     & \checkmark     & \checkmark     & \textbf{46.7} & \textbf{50.8} & \textbf{0.36} \\
    \bottomrule
    \end{tabular}%
  \label{tab_aa}%
\end{table}%

\subsection{Discussion}

In this section, we first visualize the attention maps learned by the model, then discuss 
the FID results, and finally analyze the failure cases.

\subsubsection{Attention Map Visualization}

\begin{figure}[!t]
\centering
\includegraphics[width=3.45in]{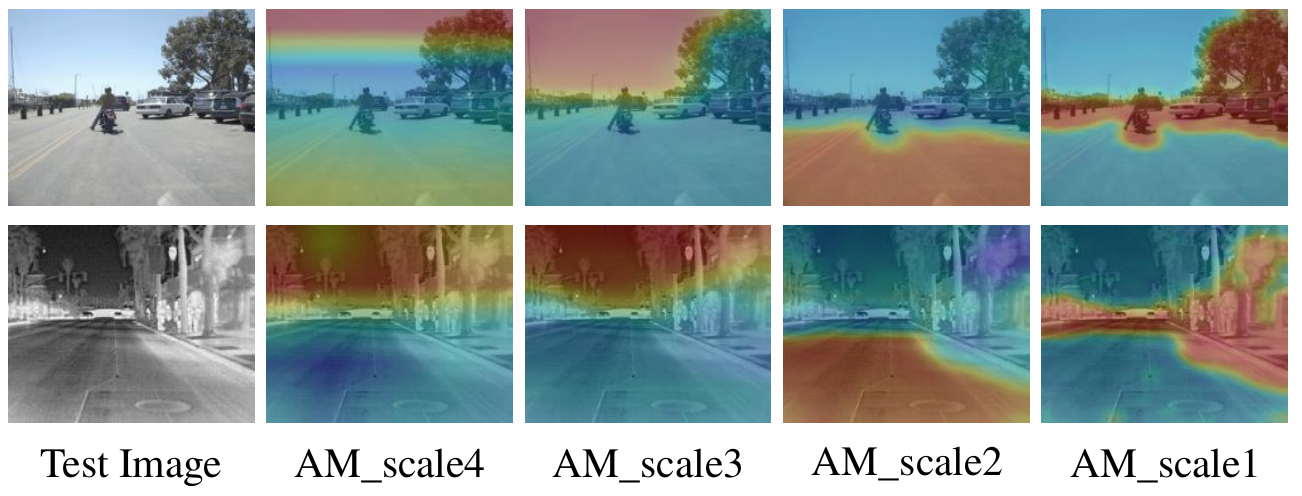}
\caption{Attention map visualization learned through TDGA module. The second to the 
fifth column indicate the attentional maps from the fourth to the first scale, respectively.}
\label{fig_att}
\end{figure}

We visualize the attention maps of a pair of DC and NTIR images from the 
test set of the FLIR dataset to see whether they can learn spatially separated attention. 
The attention maps are presented in Fig. \ref{fig_att}. As shown in the second column of the figure, 
the models all tend to pay attention to the top region of the image to capture contextual 
information regardless of whether the image is a visible or infrared image. From the third 
to the first scale of the attentional map, the model tends to pay 
attention to the top, bottom and middle regions of the image, which correspond to the sky, 
road and object-related regions, respectively. This hierarchical feature encoding manner is 
beneficial for reducing the semantic entanglement in the neighboring space by using contextual 
information. The aforementioned experiments on FLIR and KAIST datasets demonstrate the 
effectiveness of this attention pattern.

\subsubsection{FID Results}

\begin{figure}[!t]
\centering
\includegraphics[width=3.45in]{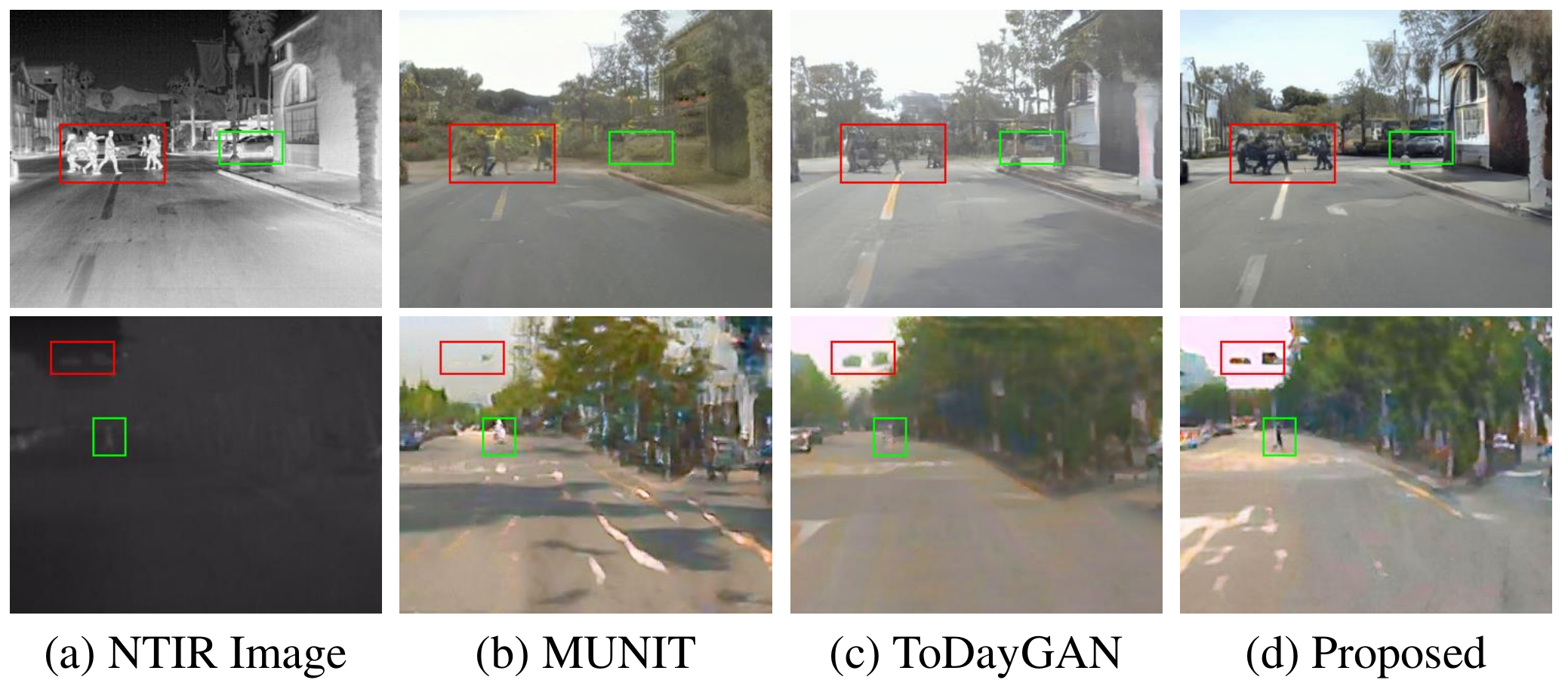}
\caption{Qualitative comparison of translation results. Areas covered by the red and green 
boxes are worth attention.}
\label{fig_fid}
\end{figure}

Due to its widespread adoption in I2I image translation performance evaluation, we report the 
FID scores of different methods for the NTIR2DC task in Table \ref{tab_fid}. As shown in the table, the 
FID score of the proposed method ranks in the middle of all compared methods, while MUNIT 
and ToDayGAN outperform PearlGAN on both datasets. However, the FID score is biased, as it is 
related to the number of samples \cite{2020-CVPR-Chong}. Moreover, the FID score only 
considers the similarity of the feature distribution of all samples and fails to measure 
differences in content and geometry before and after translation \cite{2017-NIPS-Heusel}. 
As shown in Fig. \ref{fig_fid}, MUNIT and TDG translate people into vegetation 
(i.e., the first row) or road (i.e., the second row) in order to obtain more 
realistic texture information, which 
is unacceptable for the domain adaptation scenarios, requiring rigorous preservation of 
image content. In contrast, our approach encourages the model to maintain the geometry 
and reduce the semantic encoding ambiguity of the original image, and thus may generate 
local unnatural textures, leading to degradation of the FID score. Therefore, establishing 
a metric that simultaneously balances texture naturalness, content preservation and 
geometric consistency is an imperative research direction for future unpaired I2I translation work.

\begin{table}[htbp]
  \centering
  \caption{FID Results on FLIR and KAIST Datasets}
    \begin{tabular}{ccc} \toprule
          & FLIR Dataset & KAIST Dataset \\ \hline
    CycleGAN \cite{2017-CVPR-Zhu} & 76.0  & 132.7  \\
    UNIT \cite{2017-NIPS-Liu}  & 78.1  & 91.9  \\
    MUNIT \cite{2018-ECCV-Huang} & \textbf{39.0} & 98.3  \\
    ToDayGAN \cite{2019-ICRA-Anoosheh} & 56.9  & \textbf{90.7} \\
    UGATIT \cite{2019-ICLR-Kim} & 69.8  & 99.8  \\
    DRIT++ \cite{2020-IJCV-Lee} & 58.9  & 105.9  \\
    ForkGAN \cite{2020-ECCV-Zheng} & 99.8  & 175.3  \\
    Proposed & 62.7  & 102.2  \\
    \bottomrule
    \end{tabular}%
  \label{tab_fid}%
\end{table}%

\subsubsection{Failure Cases}

\begin{figure}[!t]
\centering
\includegraphics[width=2.8in]{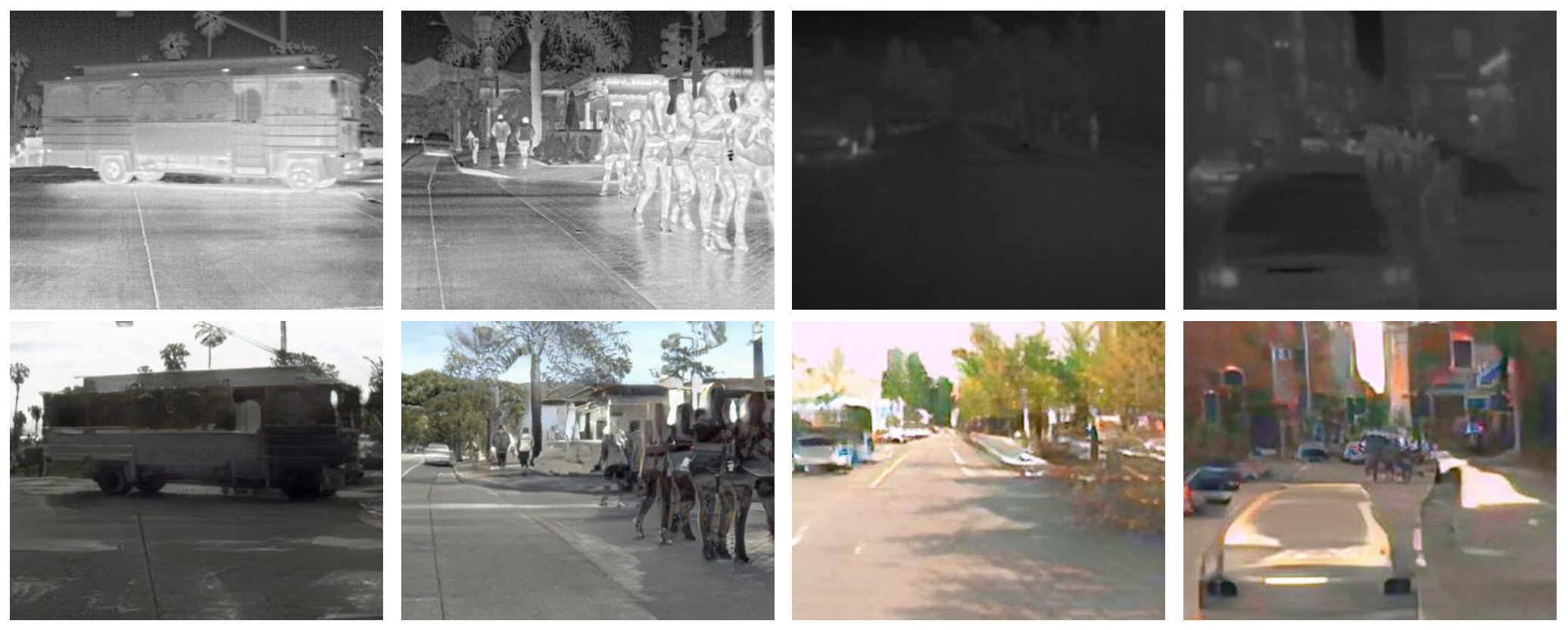}
\caption{Visualization of failure cases. The first and second rows show the NTIR images and 
their translation results, respectively.}
\label{fig_failure}
\end{figure}

Fig. \ref{fig_failure} shows some failure cases of PearlGAN, the first two and last 
two example images are from the FLIR and KAIST datasets, respectively. As shown in the 
first and third columns of the figure, the proposed method fails to generate plausible buses due to few 
training samples containing this category. However, poor translation on small sample 
categories is a common defect among all comparison methods, as shown in 
Table \ref{tab_flir_seg} and Table \ref{tab_kaist_seg}. A direct solution to this problem could be 
increasing the number of images in small sample categories by data augmentation. Although 
the proposed method can obtain natural background regions, reasonably translating objects in 
some complex and crowded scenes is still challenging, as shown in the second and fourth columns. 
To address this issue, more attempts should be made to develop better learning strategies for 
the understanding of scene layout in the NTIR images.

\section{Conclusion}

In this paper, we propose a novel framework called PearlGAN to achieve TIR image colorization, which 
is beneficial to multiple vision tasks in nighttime driving scenes. Benefitting from the 
top-down guided attention structure and elaborated attentional loss, PearlGAN can learn 
hierarchical attention to reduce the spatial entanglement of features and 
better preserve semantic information. Moreover, we propose 
a structured gradient alignment loss to encourage geometric consistency between the 
translation result and the original image. In addition, we annotate a subset of FLIR and KAIST datasets 
with pixel-wise category labels to further catalyze research on colorization 
and semantic segmentation of NTIR images. Furthermore, we introduce a new evaluation metric 
to assess the edge consistency of the translation method. Comprehensive experiments demonstrate 
the superiority of PearlGAN for semantic preservation and edge consistency in the NTIR2DC 
task. In the future, designing a more reliable image translation model to maintain semantic 
consistency is a promising direction for our further research.

\ifCLASSOPTIONcaptionsoff
  \newpage
\fi

\bibliographystyle{IEEEtran}

\bibliography{refabrv}

\end{document}